%% file: april_aigc.tex
\ifdefined\pdfminorversion
\fi

\PassOptionsToPackage{table}{xcolor}
\PassOptionsToPackage{pagebackref,breaklinks=true,colorlinks,citecolor=blue,urlcolor=blue,linkcolor=blue,bookmarks=false}{hyperref}
\PassOptionsToPackage{noabbrev,nameinlink,capitalize}{cleveref}

\documentclass[onecolumn,numbers]{april_aigc}

\input{preamble}

\usepackage[utf8]{inputenc}
\usepackage[T1]{fontenc}
\usepackage{float}
\usepackage{url}
\usepackage{graphicx}
\usepackage[percent]{overpic}
\usepackage{booktabs}
\usepackage{amsmath}
\usepackage{amssymb}
\usepackage{amsfonts}
\usepackage{microtype}
\usepackage{xspace}
\usepackage{placeins}
\usepackage{wrapfig}
\usepackage{tikz}
\usepackage{multicol}
\usepackage{multirow}
\usepackage{makecell}
\usepackage{tabularx}
\usepackage{adjustbox}
\usepackage{colortbl}
\usepackage{pifont}
\usepackage{enumitem}
\usepackage[normalem]{ulem}
\usepackage{algorithm}
\usepackage{algorithmic}
\usepackage{listings}
\usepackage{etoolbox}
\usepackage{caption}

\graphicspath{{fig/}}

\definecolor{linkcolor}{named}{aprilblue}
\definecolor{urlcolor}{RGB}{255,105,180}
\definecolor{citecolor}{RGB}{66,168,235}
\definecolor{lightgray}{rgb}{0.8, 0.8, 0.8}
\definecolor{darkgreen}{rgb}{0.00, 0.81, 0.78}
\definecolor{gray_tab}{RGB}{220, 220, 220}
\definecolor{blue_tab}{RGB}{227, 240, 251}
\definecolor{oran_tab}{RGB}{252, 242, 237}
\definecolor{whit_tab}{RGB}{255, 255, 255}
\definecolor{green_code}{RGB}{55, 126, 34}
\definecolor{oursrowblue}{RGB}{232,242,255}
\definecolor{boxteal}{RGB}{70,160,160}

\newcommand{\stardojo}{\textsc{StarDojo}}
\newcommand{\method}{\textsc{SPIKE}}
\newcommand{\samb}{\textsc{SA-MB}}

\newcommand{\best}[1]{\textbf{#1}}
\newcommand{\second}[1]{\protect\underline{#1}}
\newcommand{\std}[1]{\scalebox{0.85}{\(\pm\)#1}}

\makeatletter
\AfterEndEnvironment{algorithm}{\let\@algcomment\relax}
\AtEndEnvironment{algorithm}{\kern2pt\hrule\relax\vskip3pt\@algcomment}
\let\@algcomment\relax
\newcommand\algcomment[1]{\def\@algcomment{\footnotesize#1}}
\renewcommand\fs@ruled{\def\@fs@cfont{\bfseries}\let\@fs@capt\floatc@ruled
  \def\@fs@pre{\hrule height.8pt depth0pt \kern2pt}%
  \def\@fs@post{}%
  \def\@fs@mid{\kern2pt\hrule\kern2pt}%
  \let\@fs@iftopcapt\iftrue}
\makeatother

\DeclareCaptionFormat{custom}{{\color{aprilblue}\sffamily\textbf{#1 #2}} #3}
\titleformat*{\section}{\color{aprilblue}\Large\sffamily\bfseries}
\titleformat*{\subsection}{\color{aprilblue}\large\sffamily\bfseries}
\titleformat*{\subsubsection}{\color{aprilblue}\normalsize\sffamily\bfseries}

\usepackage{fancyhdr}
\newif\ifshowlogo
\showlogotrue
\newcommand{\insertlogo}{%
  \ifshowlogo
    \IfFileExists{assets/april_logo1.png}%
    {\includegraphics[height=0.68cm]{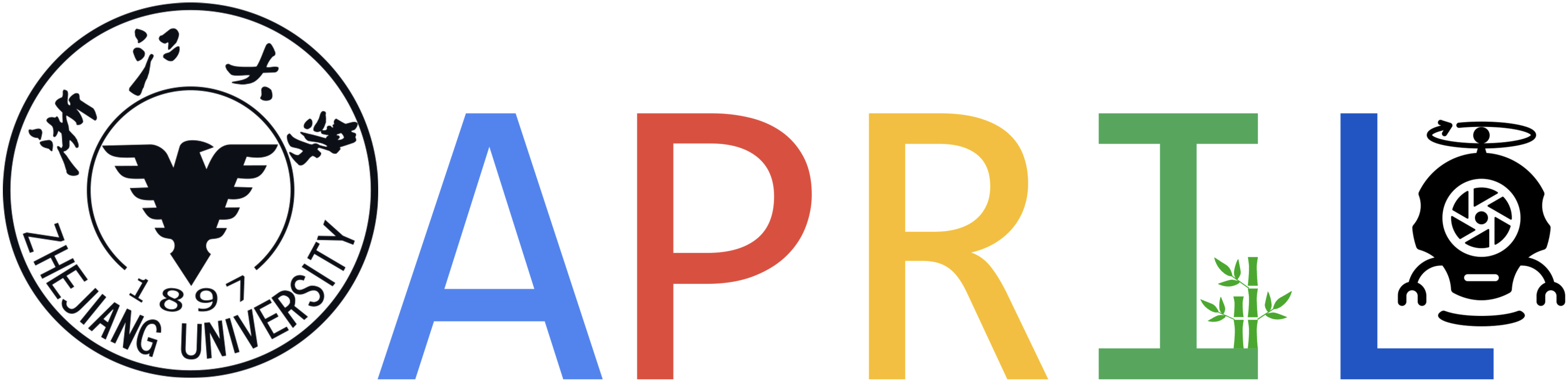}}%
    {}%
  \fi
}
\newif\ifshowtoc
\showtocfalse

\hypersetup{
  colorlinks=true,
  linkcolor=linkcolor,
  urlcolor=urlcolor,
  citecolor=citecolor,
  pdfauthor={Wencan Jiang, Jiangning Zhang, Jianbiao Mei, Jinzhuo Liu, Yu Yang, Xiaobin Hu, Zhucun Xue, Yong Liu, Dacheng Tao},
  pdftitle={SPIKE: An Adaptive Dual Controller Framework for Cost-Efficient Long-Horizon Game Agents}
}

\renewcommand{\title}[1]{\def\titlelist{{\fontsize{20pt}{28pt}\selectfont\sffamily\bfseries #1}}}
\title{SPIKE: An Adaptive Dual Controller Framework for Cost-Efficient Long-Horizon Game Agents}

\author[1,*]{Wencan Jiang}
\author[1,*\raisebox{-0.2em}{\includegraphics[height=0.85em]{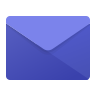}}]{Jiangning Zhang}
\author[1]{Jianbiao Mei}
\author[1]{Jinzhuo Liu}
\author[1]{Yu Yang}
\author[2]{Xiaobin Hu}
\author[1]{Zhucun Xue}
\author[1]{Yong Liu}
\author[3]{Dacheng Tao}

\affiliation[1]{Zhejiang University}
\affiliation[2]{National University of Singapore}
\affiliation[3]{Nanyang Technological University}

\contribution[*]{Equal contribution.}
\contribution[\dagger]{Corresponding author: Jiangning Zhang.}

\abstract{
Long-horizon multimodal agents in open-world games must stay goal-directed
across many low-level interactions under tight token and latency budgets.
Existing approaches often trade off costly per-step reasoning against reactive
execution that can drift, repeat failures, and recover poorly. Our key idea is
to reuse strategic reasoning across locally stable segments and reinvoke it at
event boundaries. We present \method{}, an \textbf{adaptive dual controller
framework} for cost-efficient long-horizon game control. Its Strategic
Controller performs low-frequency global planning, failure analysis, and
recovery, while its Reactive Controller handles fast local execution under a
strict token budget. An \textbf{Event Trigger} monitors visual change, task
progress, repeated actions, and failure signals to decide when control should
stay reactive or escalate to strategic reasoning. \textbf{Hierarchical Memory}
separates short-term experience reuse in the State-Action Memory Bank
(\samb{}) from structured evidence in the State-Action Knowledge Graph
(SA-KG), allowing each controller to retrieve the context it needs. This design
reuses strategic proposals over multiple reactive steps, supports local
override when plans become stale, and reserves expensive reasoning for moments
where extra deliberation is useful. On the Lite-100 split of \stardojo{},
\method{} improves Lite-100 success rate (SR) by \textbf{5.0 percentage points}
(38.5\% relative) over the strongest Lite-100 baseline and Budgeted SR by
\textbf{9.3 points} (75.6\% relative) over the strongest budgeted baseline. It
also reduces token consumption by \textbf{54.9\%} and latency by
\textbf{40.8\%}. Ablations show that \textbf{event triggering},
\textbf{reactive override}, and \textbf{heterogeneous memory} each contribute
to success and recovery, supporting \textbf{selective reasoning} rather than
reasoning at every step.
}

\coverdate{\today}
\covercorrespondence{186368@zju.edu.cn}
\coversourcecode{https://github.com/wencanjiang/SPIKE}
\coverproject{https://wencanjiang.github.io/projects/SPIKE/}

\begin{document}

\maketitle
\thispagestyle{plain}

\ifshowtoc
  \clearpage
  \setcounter{tocdepth}{2}
  \tableofcontents
  \vspace{1cm}
  \clearpage
\fi

\input{sec/1_introduction}
\input{sec/2_related_work}
\input{sec/3_method}
\input{sec/4_experiment}
\input{sec/5_conclusion}

\bibliography{april_aigc}

\clearpage
\appendix
\renewcommand\thefigure{A\arabic{figure}}
\renewcommand\thetable{A\arabic{table}}
\renewcommand\theequation{A\arabic{equation}}
\renewcommand\theHfigure{A\arabic{figure}}
\renewcommand\theHtable{A\arabic{table}}
\renewcommand\theHequation{A\arabic{equation}}
\setcounter{equation}{0}
\setcounter{table}{0}
\setcounter{figure}{0}
\input{sec/X_suppl}

\end{document}

%% file: preamble.tex
\microtypesetup{expansion=false}

%% file: sec/1_introduction.tex
\section{Introduction}
\label{sec:introduction}

Recent advances in Large Language Models (LLMs), Vision-Language Models
(VLMs), and vision-action foundation models have enabled increasingly capable
agents for open-world games~\citep{wang2024voyager,magne2026nitrogen}, web
navigation~\citep{zhou2023webarena,deng2023mind2web}, and general computer
control~\citep{liu2024agentbench,tan2024cradle}. Unlike single-step question
answering, these environments require agents to act over long, changing
interaction histories with delayed feedback. In tasks such as
\stardojo{}~\citep{tan2025stardojo}, which spans 1,000 tasks and a Lite-100
subset, an agent may need dozens or hundreds of steps to coordinate tool use,
navigate, handle menus, switch subgoals, and recover from failures without a
privileged game-state API. The central challenge is therefore \textbf{closed-loop
control under scarce deliberation}: reasoning should be spent on bottlenecks and
recovery, while routine low-level actions must stay cheap. This creates a
\textbf{planning-latency-memory trilemma}: deeper reasoning improves
deliberation but raises inference cost, lightweight reactive execution is fast
but short-sighted, and naive memory reuse introduces retrieval noise.

\begin{figure}[t!]
    \centering
    \includegraphics[width=0.98\textwidth]{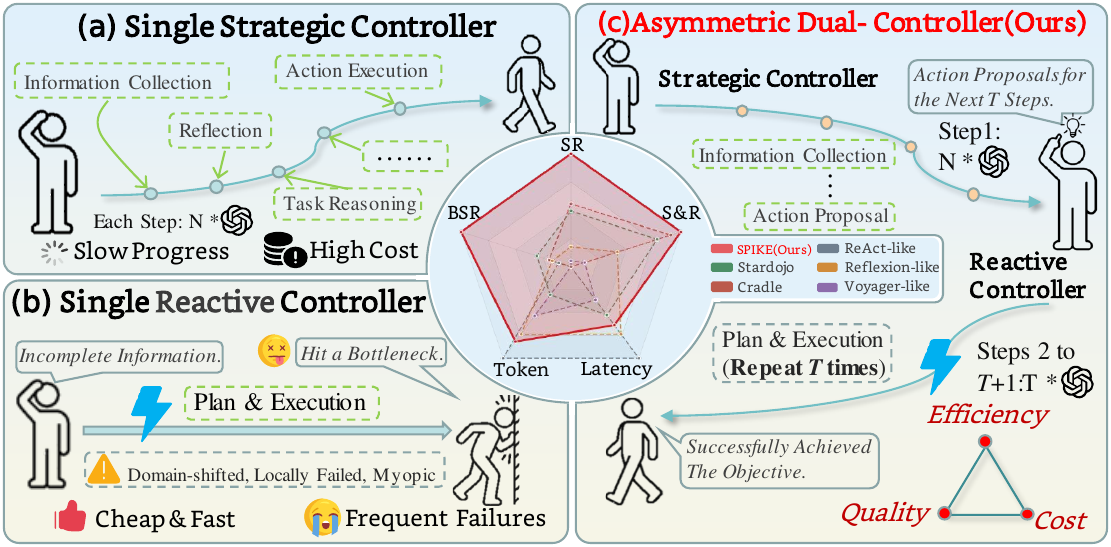}
    \caption{\textbf{Motivation for event-triggered deliberation.}
    Single-Strategic \textbf{\textit{(a)}} control reasons carefully but is expensive, while Single-
    Reactive \textbf{\textit{(b)}} control is cheap but brittle under drift and bottlenecks. \textbf{\textit{(c)}} \method{}
    allocates strategic reasoning only at likely bottlenecks, aiming for higher
    success and recovery with lower per-step cost.}
    \label{fig:motivation}
    \vspace{-0.6em}
\end{figure}

Current agents address only parts of this control problem. As illustrated in
Figure~\mbox{\ref{fig:motivation}}-(a), planning-intensive frameworks such as
CRADLE can reason effectively, but repeated calls to large models make
high-frequency control expensive in tokens and latency~\citep{tan2024cradle}.
Conversely, Figure~\mbox{\ref{fig:motivation}}-(b) captures reactive agents such
as ReAct: they execute cheaply but become unreliable under drift, execution
failure, repetition, and multi-stage transitions~\citep{yao2023react}.
The desired behavior, shown in Figure~\mbox{\ref{fig:motivation}}-(c), reuses
strategic guidance across routine reactive steps while retaining the ability to
replan when the local state changes. Fast-slow and hierarchical agents, including
SayCan, Inner Monologue, and SwiftSage~\citep{ahn2022saycan,huang2022innermonologue,lin2023swiftsage},
as well as ADaPT and Optimus-3~\citep{prasad2024adapt,li2025optimus3}, move
toward selective reasoning. However, multimodal open worlds leave three
gaps: triggering is often rigid, sudden visual changes are hard to detect from
text-only cues, and the fast layer is frequently treated as a passive executor.
Meanwhile, long-horizon memory is often a flat prompt buffer or single
retrieval pool, weakening cross-episode reuse and increasing retrieval
noise~\mbox{\citep{park2023generativeagents,packer2023memgpt}}.

Motivated by these observations, we propose \method{}, an \textbf{event-triggered
deliberation framework} for cost-efficient long-horizon game agents. Our core
insight is that strategic reasoning should be amortized across locally stable
segments and re-invoked only at event boundaries. Long-horizon game control is
temporally uneven: most steps lie in stable local execution, while a few
boundaries dominate recoverable failures. This turns reasoning into a resource
allocation problem. \method{} therefore couples three mechanisms: the Event
Trigger decides \emph{when} expensive reasoning is worth invoking; the bounded
Reactive Controller decides \emph{how} to act safely between deliberations rather
than serving as a passive executor; and controller-specific Hierarchical Memory
decides \emph{what evidence} supports local action reuse versus strategic
replanning. As detailed in Section~\ref{sec:method}, the memory separates the
State-Action Memory Bank (\samb{}) from the State-Action Knowledge Graph
(SA-KG), preventing routine execution and replanning from sharing one noisy
retrieval context.

Overall, our main contributions are:
\begin{itemize}
    \item \textbf{Event-triggered amortized deliberation.} We formulate
    long-horizon multimodal control as budgeted allocation of strategic reasoning
    over event boundaries.
    \item \textbf{Adaptive dual-controller execution.} \method{} uses strategic
    planning at escalation points and bounded reactive override during stable
    local execution.
    \item \textbf{Controller-specific hierarchical memory.} \method{} separates
    \samb{} for local state-action reuse from SA-KG for structured strategic
    evidence.
    \item \textbf{Cost-aware evidence.} On \stardojo{} Lite-100 and RDR2,
    \method{} improves success--cost trade-offs, with ablations isolating trigger
    timing, override, and memory structure.
\end{itemize}

%% file: sec/2_related_work.tex
\section{Related Work}
\label{sec:related_work}

\subsection{Computer Control and Game Environments}
\label{sec:related_work_computer_control}

Benchmarks for agents built on foundation models mainly come from two lines.
The first studies software and general computer control, including web,
mobile, desktop, and real computer tasks~\citep{liu2024agentbench,zhou2023webarena,tan2024cradle}.
Recent benchmarks such as ScenDroid and GameWorld/V-MAGE further sharpen this line~\citep{scendroid2026,ouyang2026gameworld,zheng2025vmage}.
These settings are useful for tool use and interface grounding, but they only
partly capture the navigation, scene changes, and recovery pressure of dynamic
open-world games.
ALFWorld aligns text interaction with embodied household environments, while
MineDojo provides an open-ended Minecraft benchmark with internet-scale
knowledge~\citep{shridhar2021alfworld,fan2022minedojo}. Voyager shows that
large models can support long-running exploration in Minecraft-like
worlds~\citep{wang2024voyager}. TextAtari extends this direction to 100K-frame
decision-making and makes the trade-off between expensive reasoning and
reactive execution explicit~\citep{textatari2025}.
\stardojo{}~\citep{tan2025stardojo} lies between these traditions: it keeps
open-world dynamics but requires control without a privileged game-state API,
using screen-derived observations and a shared executable action interface,
making low-level reliability and control scheduling central.

\subsection{LLM/VLM Agents for Computer Tasks and Games}
\label{sec:related_work_llm_vlm_agents}

LLM and VLM agents have made strong progress on multi-step computer
interaction. ReAct established the observe-reason-act loop~\citep{yao2023react}.
Later agents extended this idea to web navigation, app use, and general
computer control~\citep{zhou2023webarena,deng2023mind2web,tan2024cradle}.
In games, Voyager demonstrates strong large-model planning, while TextAtari
shows how long-horizon tasks expose the tension between costly reasoning and
reactive execution~\citep{wang2024voyager,textatari2025}.
DEPS, Lumine, and SIMA extend this line with open-world control~\citep{wang2023deps,tan2025lumine,abiraad2024scaling},
while JARVIS-1 adds memory-augmented multimodal control~\citep{wang2023jarvis1}.

\paragraph{Scope of comparison.}
Recent open-world game-agent systems and benchmarks are important context but
are not always direct same-table baselines for \method{}. NitroGen trains a
vision-action foundation model via large-scale behavior cloning on 40,000 hours
of gameplay across more than 1,000 games~\citep{magne2026nitrogen}, whereas
\method{} studies test-time controller allocation without additional agent
training. Lumine targets real-time 3D open-world control with an end-to-end VLM
agent~\citep{tan2025lumine}, while \stardojo{} Lite-100 uses benchmark tasks,
shared runner feedback, and a fixed executable action interface. GameWorld
standardizes verifiable evaluation across 34 browser games and 170 tasks, and
V-MAGE provides a vision-centric protocol~\citep{ouyang2026gameworld,zheng2025vmage}.
Because these works differ in training, environments, action interfaces, and
verification signals, we discuss them as related positioning rather than direct
baselines.

However, many game agents still rely on privileged observations, task
scaffolding, or in-environment training. \method{} instead focuses on a setting
with no training and no privileged game-state API: the Reactive Controller must
remain responsive under screen-grounded interaction, while strategic reasoning,
used only when needed, handles drift, recovery, and task-level redirection.

\subsection{Dual Controller Frameworks and Memory for Long-Horizon Agents}
\label{sec:related_work_dual_controller_memory}

Fast-slow and hierarchical control manage the trade-off between costly reasoning
and real-time execution. SayCan combines language planning with affordances, Inner Monologue
uses closed-loop language feedback, and SwiftSage explores fast-slow
cooperation~\citep{ahn2022saycan,huang2022innermonologue,lin2023swiftsage}.
ADaPT and AdaPlanner further study on-demand decomposition and
feedback-adaptive planning~\citep{prasad2024adapt,sun2023adaplanner}.
Search methods, including Tree of Thoughts, RAP, and LATS, show the value of
deliberate exploration~\citep{yao2023treeofthoughts,hao2023rap,zhou2024lats}.
Reflexion further highlights feedback and retry for recovery~\citep{shinn2023reflexion}.
Many prior agents remain planning-intensive or rely on fixed triggering. In
contrast, \method{} uses multimodal event triggering, local autonomous override,
and graded recovery so that strategic reasoning is called only when needed.

Memory is equally important for long-horizon behavior. Generative Agents,
Reflexion, and MemGPT show that retrieval and reflection can help agents go
beyond a single prompt window~\citep{park2023generativeagents,shinn2023reflexion,packer2023memgpt}.
HiAgent and Mem0 further study hierarchical working memory and scalable
long-term memory~\citep{hu2025hiagent,chhikara2025mem0}. Retrieval-Augmented
Generation (RAG), GraphRAG,
and HippoRAG show how external stores and graph retrieval can extend model
context~\citep{lewis2020rag,edge2024graphrag,gutierrez2024hipporag}.
Recent agents without privileged environment APIs also use state-action graphs
for exploration and delayed reward planning~\citep{tang2025experiencedriven}.
AgentOdyssey studies open-ended text games with test-time continual learning,
highlighting exploration, episodic memory, and planning cost~\citep{agentodyssey2026}.
Our memory design follows this direction but separates roles: \samb{} supports
fast local reuse, whereas SA-KG provides evidence for strategic checking and
replanning.

%% file: sec/3_method.tex
\section{The SPIKE Framework}
\label{sec:method}

% 中文：我们将长时程开放世界控制建模为一个包含多模态观测、分层记忆、显式恢复和 token--latency 预算的序列决策问题。
% 在每一步，智能体根据当前视觉场景、文本上下文、推断的 UI 状态、近期动作和可选执行日志选择可执行动作。
We model long-horizon open-world control as sequential decision making with
multimodal observations, layered memory, recovery, and token--latency budgets:
the Event Trigger decides when to deliberate, the Reactive Controller acts
between deliberations, and Hierarchical Memory retrieves evidence at routine and
escalation boundaries.

Open-world control has a cost-value mismatch: routine steps need fast execution,
whereas rare scene changes, stalls, and failures require expensive reasoning.
Always-strategic agents spend large-model calls on routine motion; purely
reactive agents lack reliable triggers for stale plans. \method{} treats
strategic reasoning as event-triggered, amortizes proposals over stable segments,
and retrieves different evidence for local and escalated steps.

\begin{figure}[t!]
    \centering
    \includegraphics[width=0.98\linewidth]{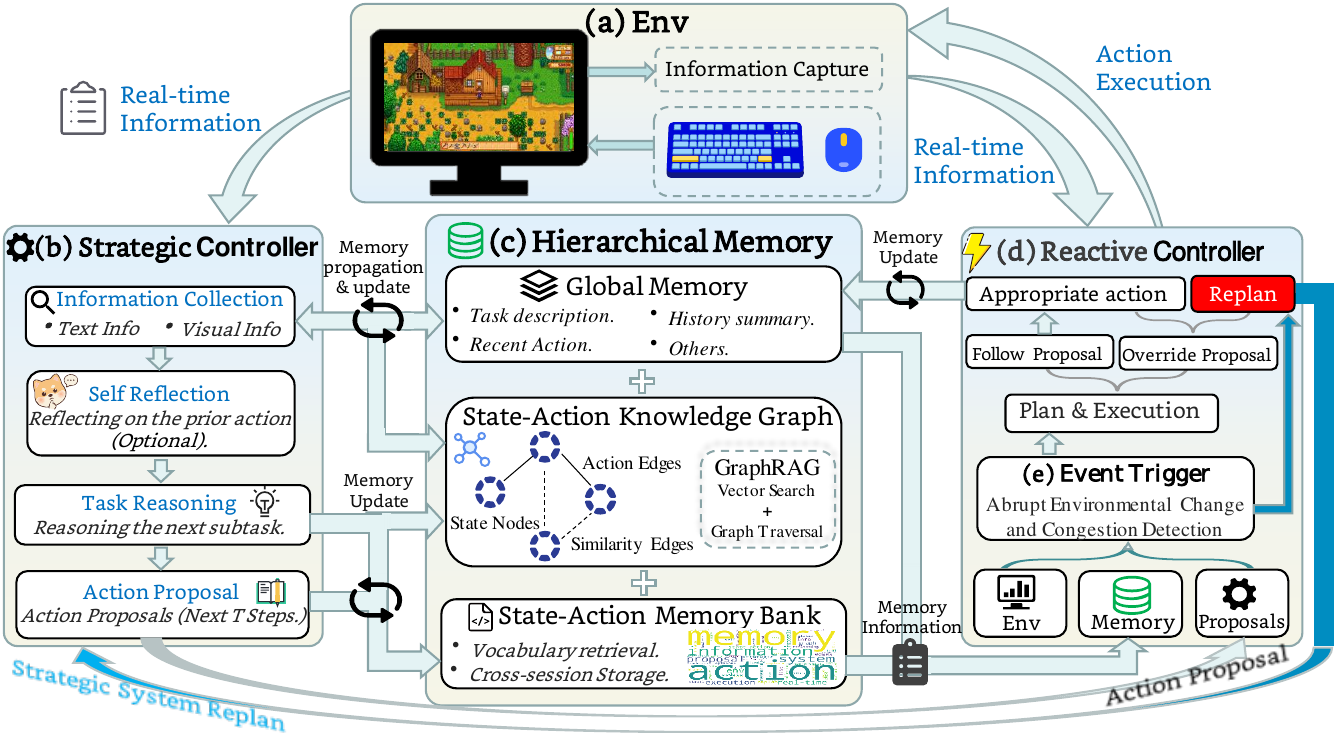}
    \caption{\textbf{Closed-loop event-triggered deliberation in \method{}.} The Strategic Controller performs information collection,
    optional Self Reflection, task reasoning, and short-horizon action proposal;
    the Reactive Controller uses real-time signals and retrieved memory to
    follow, override, or trigger replanning. The Hierarchical Memory module
    combines Global Memory, the State-Action Memory Bank (\samb{}), and the
    State-Action Knowledge Graph (SA-KG) through vector search and graph
    traversal, then propagates updates after action execution.}
    \label{fig:overview}
    \vspace{-0.5em}
\end{figure}

\subsection{Event-Triggered Dual Controller Framework}
\label{subsec:overview}

% 中文：受双过程认知理论启发，\method{} 采用非对称双控制器设计。
% \strategicsystem{} 负责低频的全局规划、失败分析和任务级重定向；\reactivesystem{} 在严格 token 预算下负责低延迟局部执行。
% 两个控制器由 Event Trigger 协调；Hierarchical Memory 为不同控制器提供各自所需的上下文。
For an episode with horizon $H$, let
$o_t=(v_t,u_t,f_t)$ denote the observation at step $t$, where $v_t$ is the
visual screenshot, $u_t$ is textual or user interface (UI) context, and $f_t$ is runner feedback
such as progress, position, inventory, or dialogue changes. The agent chooses an
executable game action $a_t\in\mathcal{A}$ from history
$h_t=(o_1,a_1,\ldots,o_t)$. An event-trigger decision $g_t\in\{0,1\}$ selects the
Strategic Controller ($g_t=1$) or Reactive Controller ($g_t=0$). For trajectory
$\tau=(o_1,a_1,\ldots,o_H)$, success is $S(\tau)\in\{0,1\}$ and the deliberation
budget is $B$. Let $C_t(g_t,h_t)$ denote the history-dependent model-call, token,
or latency cost. The target is
\begin{equation}
\label{eq:budgeted_control_objective}
\max_{\pi_R,\pi_S,g}\mathbb{E}[S(\tau)]
\quad
\mathrm{s.t.}\quad
\mathbb{E}\!\left[\sum_{t=1}^{H} C_t(g_t,h_t)\right]\le B .
\end{equation}
Inspired by dual-process theories of cognition \citep{kahneman2011thinking},
\method{} separates slow, high-cost planning from fast local execution. The
Strategic Controller handles planning, failure analysis, and task redirection;
the Reactive Controller executes local actions under a strict token budget; the
Event Trigger and Hierarchical Memory allocate deliberation and evidence.
This differs from always-deliberate agents, which call the expensive planner at
nearly every step, from fixed fast-slow pipelines, which switch controllers by a
static schedule, and from memory-augmented agents with a single retrieval pool.
In \method{}, controller allocation, local override, and memory retrieval are
all conditioned on detected event boundaries, so the framework changes both
\emph{when} reasoning is invoked and \emph{what} evidence each controller sees.

% 中文：如图~\mbox{\ref{fig:overview}} 所示，\method{} 包含 Env and Information Capture、Strategic Controller、Hierarchical Memory、Reactive Controller 和 Event Trigger 五个模块。
% 系统遵循统一闭环：信息捕获、路由、记忆检索、动作提议、动作执行与记忆更新。
% Env and Information Capture 提供实时视觉和文本信息并刷新 Hierarchical Memory；Event Trigger 检查场景变化、卡死状态和进展信号。
% 常规状态由 \reactivesystem{} 根据 State-Action Memory Bank 中的 Memory Information 和当前子任务直接处理。
% 当检测到突发视觉变化、重复失败或进展不足时，Event Trigger 将控制权升级给 \strategicsystem{}，生成未来 $T$ 步的短程计划。
% 随后 \reactivesystem{} 可以遵循该计划、在局部证据表明计划失效时覆盖它，或请求重规划；动作执行后，Memory Update 将结果写回局部和长期记忆。

As illustrated in Figure~\mbox{\ref{fig:overview}}, \method{} consists of Env
and Information Capture, Strategic Controller, Hierarchical Memory, Reactive
Controller, and Event Trigger. Together they form a closed loop:
capture produces $o_t$ and updates $h_t$; the Event Trigger scores whether the
current history marks an event boundary; the selected controller retrieves
controller-specific memory and emits $a_t$; execution feedback updates
\samb{}, SA-KG, and $h_{t+1}$; the next step repeats this trigger--act--update
cycle.
\vspace{-0.35em}

\begin{figure}[t!]
    \centering
    \includegraphics[width=0.94\linewidth]{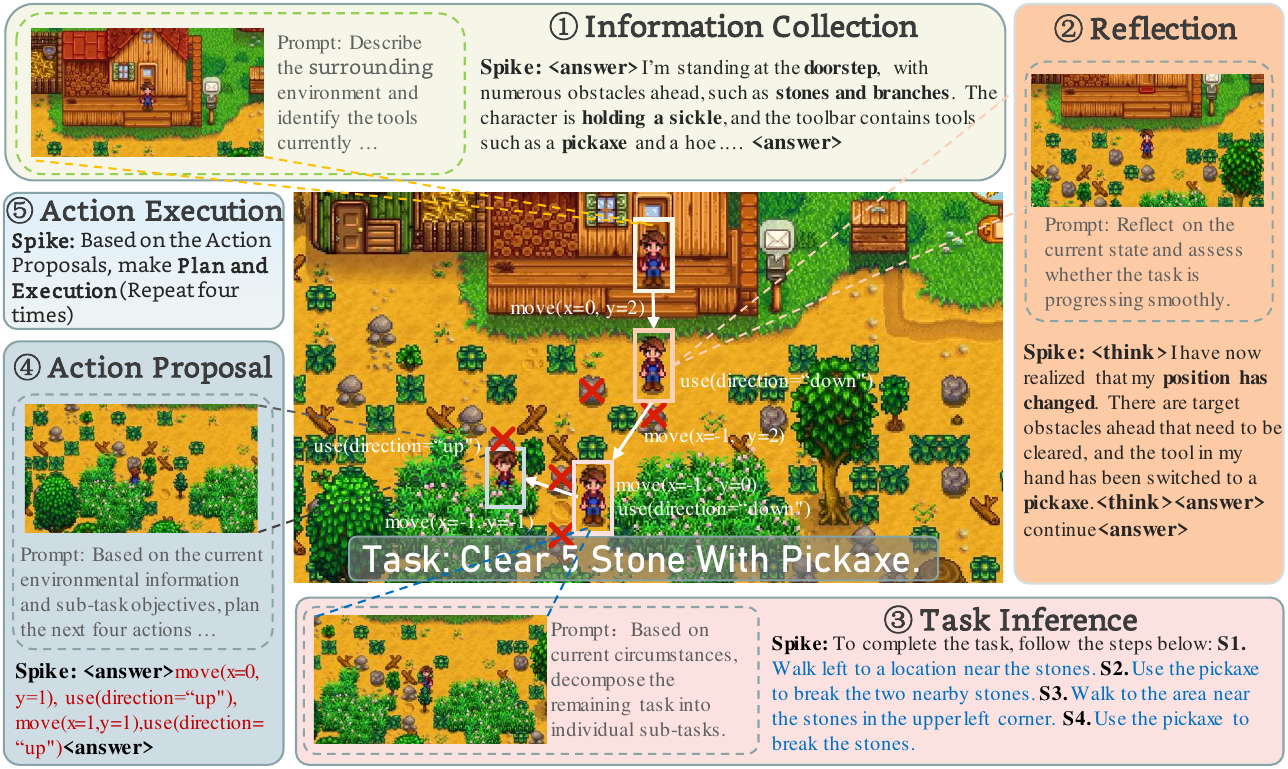}
    \caption{\textbf{Strategic Controller workflow in one escalated planning cycle.}
    The example shows how \method{} collects scene evidence, checks progress,
    reasons over the task, and emits a short plan for Reactive Controller execution.}
    \label{fig:workflow}
    \vspace{-0.5em}
\end{figure}

\paragraph{Controller roles.}
The Strategic Controller is activated when periodic refresh, stuck-state, or
recovery signals require deliberate reasoning. It integrates the current
observation, task summary, recent failure traces, and structured long-term
memory, then follows the workflow
\begin{center}
\begingroup
\vspace{-0.35em}

\begin{tikzpicture}
\node[
    draw=boxteal,
    fill=oursrowblue,
    rounded corners=5pt,
    line width=0.8pt,
    inner xsep=6pt,
    inner ysep=4pt,
    text width=0.88\linewidth,
    align=center
] {Information Collection -> Self Reflection -> Task Reasoning -> Action Proposal.};
\end{tikzpicture}
\vspace{-0.35em}

\endgroup
\end{center}
Figure~\mbox{\ref{fig:workflow}} expands the Strategic Controller workflow. The
optional Self Reflection stage handles unclear evidence, contradiction, or
repeated failure. The output is a short-horizon proposal for the next $T$ steps,
long enough to guide local execution and compact enough to revise after a scene
change.

The Reactive Controller handles routine steps under a strict token budget. It
uses the latest strategic proposal, current observation, and Memory Information
from \samb{} to emit local actions. When new evidence invalidates the proposal,
it may only apply bounded local corrections such as avoidance, repositioning, or
tool reselection; it cannot change the active subgoal, and failed corrections
return control to the Event Trigger.

\subsection{Event Trigger as an Online Surrogate for Budgeted Deliberation}
\label{subsec:controller_scheduler}

The Event Trigger operationalizes the budgeted objective as a practical online
event-boundary decision. Ideally, each step would choose whether the expected
utility of strategic deliberation justifies its extra cost:
\vspace{-0.35em}
\begin{equation}
\label{eq:event_trigger_optimization}
g_t^\star=\arg\max_{g\in\{0,1\}}
\widehat{U}(g\mid h_t)-\lambda C_t(g,h_t),
\end{equation}
\vspace{-0.1em}
where $\widehat{U}(g\mid h_t)$ estimates progress and recovery value from
history and $\lambda$ controls the cost penalty. Because the true value of
deliberation is not observable online from sparse task success, \method{} uses a
reproducible surrogate event detector. Eq.~\mbox{\ref{eq:scheduler_routing}} uses
fixed thresholds over visual discontinuity, progress stagnation, behavioral
repetition, and execution failure; these signals identify stale strategic
proposals or unreliable local execution rather than estimating task success.
Appendix Sec.~\mbox{\ref{sec:appendix_event_algorithm}} gives the trigger--act--update loop.

The trigger variable $g_t \in \{0,1\}$ indicates whether step $t$ escalates:
\vspace{-0.35em}
\begin{equation}
\label{eq:scheduler_routing}
g_t=\mathbb{I}\!\left[(c_t\ge T)\vee(d_t>\tau_v)\vee(z_t\ge\tau_z)\vee(r_t\ge\tau_r\wedge z_t\ge\tau_{rz})\vee(\ell_t\ge\tau_\ell)\right].
\end{equation}
where $c_t$ counts Reactive steps since the last Strategic call and resets after
it, $T$ is the refresh interval, and
$\tau_v,\tau_z,\tau_r,\tau_{rz},\tau_\ell$ are fixed triggering thresholds.
The detector maps signals to observable failure modes: $d_t=1-\cos(\phi(v_t),\phi(v_{t-1}))$
catches stale plans after visual discontinuities, $z_t$ catches stalled routines,
$r_t$ catches behavior loops within the last $W$ actions, and
$\ell_t\in\{0,1,2,3\}$ catches execution failure. We set
$\Delta_t=q_t-q_{t-1}$, where $q_t$ is a deterministic progress counter derived
from the shared \stardojo{} runner feedback, not a privileged game-state API;
$z_t=0$ when $\Delta_t>0$ and $z_t=z_{t-1}+1$ otherwise. All baselines receive
the same raw observation, action schema, execution feedback, and history fields;
\method{} only adds Event Trigger bookkeeping.

The fixed thresholds instantiate the event detector rather than a per-task
policy; Table~\mbox{\ref{tab:scheduler_sensitivity}} reports their local
sensitivity, and Appendix Sec.~\mbox{\ref{sec:appendix_scheduler_details}}
details their computation. These thresholds are tied to event types rather than
individual tasks: within the same game family, where camera motion, action
granularity, and progress feedback have similar scales, the same settings can be
reused without task-by-task replacement. When moving to a different game type,
the Event Trigger should be recalibrated according to the new event statistics,
such as visual dynamics, interaction pace, stall duration, and failure-message
density.
Failure levels scope recovery: F1 keeps bounded retry local, while F2/F3
escalate invalid actions, stalls, or repeated compound failures when local
context becomes unreliable.

\FloatBarrier
\subsection{Hierarchical Memory}
\label{subsec:memory}

% 中文：3.3 只保留记忆层的核心主线和两个检索打分公式，字段、融合和写回细节放到附录。
Long-horizon tasks risk context overflow, hallucination, and forgetting.
Figure~\mbox{\ref{fig:overview}}(c) organizes Hierarchical Memory around
controller-specific roles: Global Memory preserves task continuity, the
State-Action Memory Bank (\samb{}) is a low-latency local action prior for
Reactive execution, and the State-Action Knowledge Graph (SA-KG) is a strategic
evidence graph for checking and replanning rather than a flat retrieval store.

Global Memory keeps a fixed-size window of recent observations and actions,
together with the active subgoal, progress indicators, and failure summaries.
When the Event Trigger keeps execution local, the Reactive Controller
retrieves nearby state-action traces from \samb{} as local hints, with the risk
of stale habits reduced by recency, reward, and success terms. Given the current
query $x_t$, \samb{} ranks item $i$ by
\begin{equation}
\label{eq:samb_score}
S_{\mathrm{MB}}(i \mid x_t)=\rho_i(\alpha_C C_i+\alpha_L L_i+\alpha_R R_i+\alpha_P P_i).
\end{equation}
All terms are normalized before weighting: $L_i$ is the Jaccard overlap between
lowercased query and memory token sets, $C_i$ is token-frequency cosine
similarity, $R_i=(\mathrm{clip}(\mathrm{rewardEMA}_i,-1,1)+1)/2$ uses an
exponential moving average (EMA) of observed rewards, $P_i$ is
successes over attempts, $\rho_i=\exp(-\mathrm{ageHours}_i/24)$, and
$\alpha_C,\alpha_L,\alpha_R,\alpha_P$ are fixed retrieval weights. The Reactive
Controller uses the top retrieved summaries as hints rather than executable
commands.

When the Event Trigger escalates control, the Strategic Controller
consults SA-KG for plan checking rather than local traces alone.
SA-KG stores validated transitions as state nodes, action edges, and similarity
edges, inspired by state-action-result graphs in experience-driven agents
without privileged environment APIs~\citep{tang2025experiencedriven} but using
retrieved graph fragments as evidence rather than directly executing retrieved
actions.
Starting from the nearest state nodes for the current subgoal, SA-KG retrieves
graph fragments by vector search, keeps matches above $\tau_{\mathrm{KG}}$, and
ranks fragment $j$ by
\begin{equation}
\label{eq:sakg_score}
S_{\mathrm{KG}}(j \mid x_t)=\beta_C C_j+\beta_P P_j,
\end{equation}
where $C_j=1-\mathrm{dist}(x_t,s_j)$ is vector similarity and $P_j$ is the
success rate of the best outgoing action, with execution count used to break
ties. Fusion first min--max normalizes $S_{\mathrm{MB}}$ and $S_{\mathrm{KG}}$
within the current candidate sets, then uses
$S_{\mathrm{fuse}}=\lambda_{\mathrm{MB}}\tilde S_{\mathrm{MB}}+\lambda_{\mathrm{KG}}\tilde S_{\mathrm{KG}}$;
unmatched graph actions receive zero normalized SA-KG score. After execution,
\method{} writes only successful execution, interpretable recovery, or
planner-validated progress, preventing transient errors from becoming long-term
evidence. Appendix Sec.~\mbox{\ref{sec:appendix_memory_details}} gives the
storage fields, fusion score, fixed weights, and promotion rules.

%% file: sec/4_experiment.tex
\section{Experiments}
\label{sec:experiments}
\begingroup
\setlength{\textfloatsep}{8pt plus 2pt minus 2pt}
\setlength{\floatsep}{6pt plus 2pt minus 2pt}
\setlength{\intextsep}{6pt plus 2pt minus 2pt}

This section evaluates \method{} in long-horizon game control. The main
Lite-100 study, derived from \stardojo{}~\citep{tan2025stardojo}, measures
success, LLM-call efficiency, latency, and recovery, then uses task breakdowns,
Pareto analysis, ablations, and qualitative traces to explain the gains. We also
test the same adaptive dual controller on Red Dead Redemption 2 (RDR2) to assess
transfer across visual domains, action spaces, and task languages.

\subsection{Experimental Setup}

\begin{wraptable}[9]{r}{0.44\linewidth}
    \vspace{-1em}
    \centering
    \scriptsize
    \caption{\textbf{Lite-100 task statistics.}}
    \label{tab:lite100_split}
    \resizebox{\linewidth}{!}{%
    \begin{tabular}{lrrrr}
        \toprule
        Category & Total & Easy & Medium & Hard \\
        \midrule
        Farming & 21 & 14 & 3 & 4 \\
        Crafting & 14 & 7 & 4 & 3 \\
        Exploration & 28 & 15 & 8 & 5 \\
        Combat & 12 & 3 & 6 & 3 \\
        Social & 25 & 17 & 6 & 2 \\
        \midrule
        Total & 100 & 56 & 23 & 21 \\
        \bottomrule
    \end{tabular}
    }
    \vspace{-2em}
\end{wraptable}

Our main evaluation uses Lite-100 in \stardojo{}~\citep{tan2025stardojo}, which
contains 100 tasks across farming, crafting, exploration, combat, and social
interaction. Table~\mbox{\ref{tab:lite100_split}} shows the category and
difficulty split. We compare against ReAct-like~\citep{yao2023react},
Reflexion-like~\citep{shinn2023reflexion}, Voyager-like~\citep{wang2024voyager},
CRADLE~\citep{tan2024cradle}, and the benchmark-native \stardojo{} baseline.
Unless otherwise specified, all baselines and the default \method{} use the same
Qwen3.5-397B-A17B backend, environment interface, action space, prompt upper
bound, decoding settings, and, when applicable, the same frozen visual-change
encoder/descriptor. Appendix
Sec.~\mbox{\ref{sec:appendix_protocol_details}} specifies the baseline prompt
fields, history truncation, memory isolation, retry handling, and adaptation
rules.

\input{table/main_results}

We report Lite-100 success rate (SR), Budgeted SR, tokens per task, latency per step,
and Recovery/Stuck Ratio. Lite-100 SR follows the benchmark step-cap protocol,
whereas Budgeted SR uses equal difficulty-dependent LLM-call budgets
(120/200/600 calls for easy/medium/hard tasks), where one call denotes one
large-model invocation. Because this protocol constrains LLM calls rather than
environment steps, Budgeted SR should be read as an efficiency metric rather
than a stricter subset of Lite-100 SR. Unless otherwise specified, all
quantitative results report mean and standard deviation over three repeated runs. Appendix
Sec.~\mbox{\ref{sec:appendix_reproducibility}} defines call counting, and
Table~\mbox{\ref{tab:raw_recovery_events}} reports the raw events behind the
Recovery/Stuck Ratio.

\subsection{Mechanistic and Qualitative Analysis}

\begin{figure}[t!]
    \vspace{-2em}
    \centering
    \includegraphics[width=0.9\linewidth]{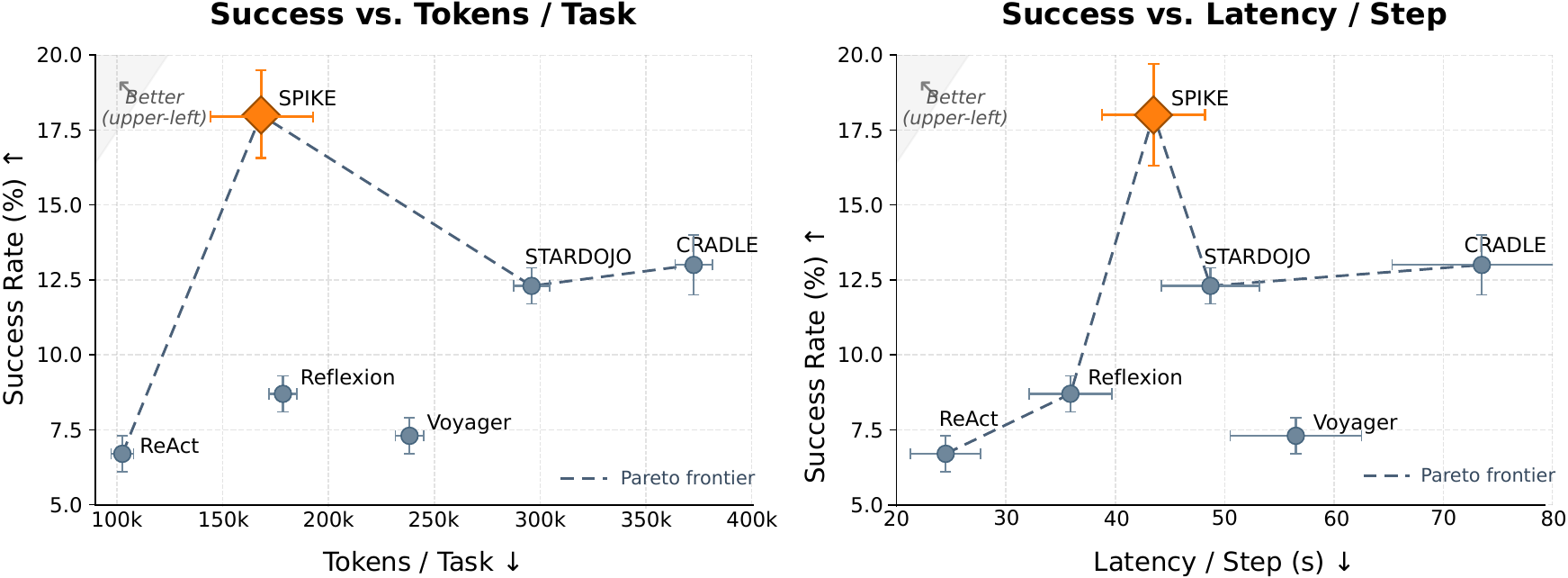}
    \caption{\textbf{Effectiveness-efficiency trade-off on Lite-100.}
    The two panels compare success against token use and latency; upper-left is better.
    \method{} stays on the Pareto frontier, improving success without simply spending more compute.}
    \label{fig:pareto_tradeoff}
    \vspace{-0.5em}
\end{figure}

\subsection{Main Results on Lite-100}

Under the same backend and environment interface,
Table~\mbox{\ref{tab:main_results}} shows that \method{} reaches the \textbf{highest
Lite-100 SR} at 18.0\% and the \textbf{highest Budgeted SR} at 21.6\%. Compared with
CRADLE, the strongest Lite-100 baseline, \method{} improves success by 5.0
points while reducing tokens from 372.5k to 168.1k and latency from 73.5\,s to
43.5\,s. It also improves Budgeted SR by 9.3 points over \stardojo{}, the
strongest budgeted baseline, showing a better success-cost trade-off rather than
a solved benchmark. This matters because Lite-100 rewards sustained progress
under fixed interaction budgets, so reducing unnecessary large-model calls can
improve both efficiency and budgeted completion.

Figure~\mbox{\ref{fig:pareto_tradeoff}} and
Table~\mbox{\ref{tab:task_breakdown}} show that this gain \textbf{does not come from
uniformly spending more compute} or from one task family: \method{} stays on the
Pareto frontier and has the highest mean success rate across all task families.
Combat and Social remain hardest due to moving enemies and long inter-house
routes; Appendix Sec.~\mbox{\ref{sec:appendix_task_family_disparities}} details
these failures. The GPT-5.4/Gemini-3.1-pro rows are controlled auxiliary backend
runs rather than isolated model-quality comparisons (Appendix
Sec.~\mbox{\ref{sec:appendix_model_token_note}}).
This consistency makes the ablations focus on where the gains arise, rather than whether they are task-specific.

\input{table/task_breakdown}

\subsection{Ablation Study}
\label{sec:ablation}

The ablation study separates controller allocation from memory design. In
Table~\mbox{\ref{tab:ablation_results}}, \emph{Single Strategic Controller} variants
retain deliberate reasoning but pay high cost, reaching at best 16.3\% SR with
305.4k tokens per task. \emph{Single Reactive Controller} variants are cheaper
but reach only 10.3\% SR even with Hierarchical Memory. The full \method{} gains
from assigning deliberate reasoning to discontinuities and cheap execution to
stable local progress. This pattern indicates that the benefit is not merely
having two controllers available, but routing control to the one whose latency
and reasoning depth match the current segment.

The component removals show that this is not just a two-controller ensemble.
Removing the \emph{Event Trigger} or \emph{reactive override} drops SR to 15.7\%,
so both escalation timing and local correction matter. Memory ablations show the
other half of the mechanism: replacing Hierarchical Memory with \emph{Global
Memory} lowers SR from 18.0\% to 12.3\%, while removing \emph{SA-MB} or
\emph{SA-KG} weakens recent-experience reuse or structured evidence. The gain
therefore comes from coupling Event Trigger decisions with controller-specific
memory, not from adding more prompt context.
\input{table/scheduler_sensitivity}

Table~\mbox{\ref{tab:scheduler_sensitivity}} tests Event Trigger sensitivity.
The default setting gives the best success--cost balance; earlier escalation
adds Strategic Controller calls but lowers SR, while delayed escalation and
removing periodic refresh cause larger drops. Appendix
Sec.~\mbox{\ref{sec:appendix_scheduler_details}} gives threshold details.

\input{table/ablation_results}

\begin{figure}[t!]
    \centering
    \includegraphics[width=1.0\linewidth]{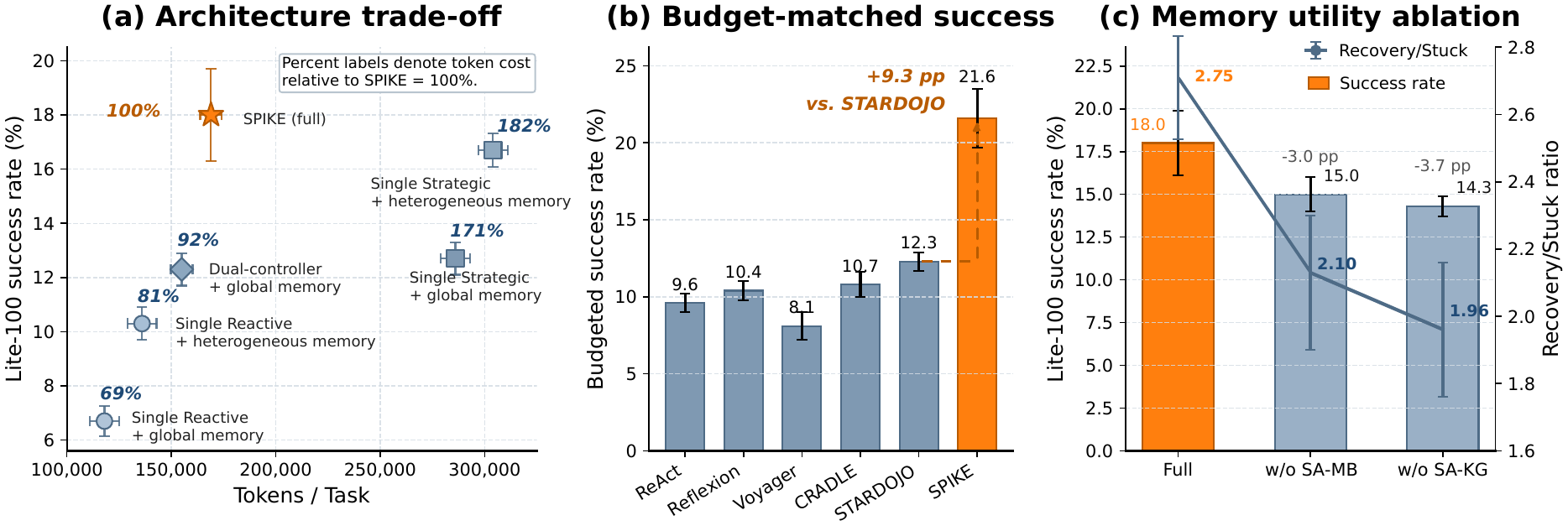}
    \vspace{-1.0em}
    \caption{\textbf{Mechanistic analysis of \method{}.} \textbf{\textit{(a)}} compares controller
    variants by SR and tokens, showing dual control improves success at lower
    cost. \textbf{\textit{(b)}} reports Budgeted
    SR under difficulty-dependent LLM-call budgets, testing whether event triggering
    delays budget exhaustion. \textbf{\textit{(c)}} removes the two memory tracks and
    reports success and Recovery/Stuck Ratio.}
    \label{fig:analysis_breakdown}
\end{figure}

\begin{wrapfigure}[21]{r}{0.70\linewidth}
    \vspace{1em}
    \centering
    \includegraphics[width=\linewidth]{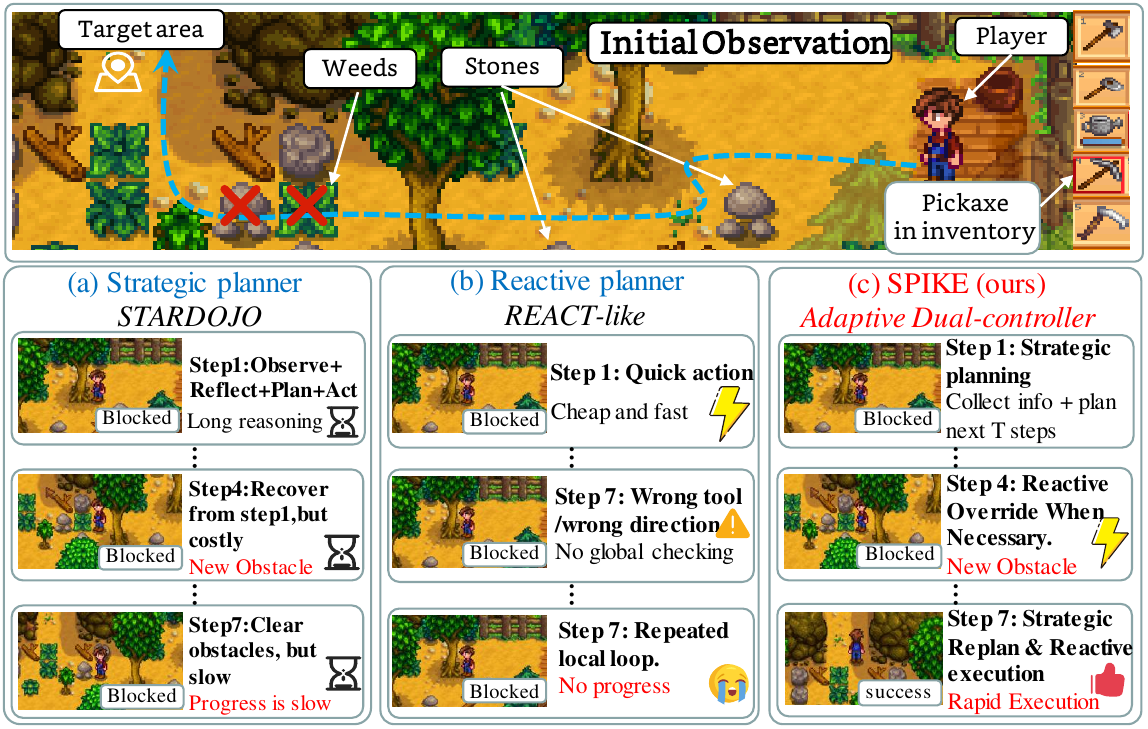}
    \caption{\textbf{Qualitative goal-reaching comparison.}
    \method{} escalates and returns to low-cost local
    execution once the plan is stable.}
    \label{fig:qualitative_analysis}
    \vspace{0.5em}
\end{wrapfigure}

Figure~\mbox{\ref{fig:analysis_breakdown}} links success and cost gains to
controller allocation: \textbf{(a)} shows that \method{} avoids both low-success
purely reactive execution and high-cost always-strategic execution. \textbf{(b)}
shows that the Event Trigger reserves Strategic Controller calls for refresh,
stuck, and recovery events while routine steps stay reactive. \textbf{(c)}
connects recovery to memory: \samb{} supports fast local reuse, while SA-KG
supplies structured evidence for checking and replanning.

Figure~\mbox{\ref{fig:qualitative_analysis}} illustrates the same mechanism during
execution. ReAct-like control keeps issuing low-cost but unproductive actions
after the visual state changes, while an always-strategic controller recovers but
continues expensive reasoning after the route pattern is understood. \method{}
separates recovery from routine execution: the Event Trigger escalates, the
Strategic Controller revises the plan with current observation and memory, and
the Reactive Controller handles local clearing.

\input{table/raw_recovery_events}

\subsection{Cross-Benchmark Generalization}

\input{table/generalization_results}

We further evaluate transfer on Red Dead Redemption 2 (RDR2), which changes the
visual domain, action space, interaction pace, and task language relative to
Lite-100. This setting is a stress test for event-triggered deliberation rather
than a benchmark-specific tuning exercise: the tasks include navigation,
following, search, protection, and combat-like objectives, while photorealistic
views and moving characters produce different visual-change and failure
statistics. The experiment follows the CRADLE benchmark setup, using the same
environment configuration, task interface, action space, and evaluation protocol
for all methods. The evaluation contains 13 selected long-horizon tasks, each
run three times; Appendix Table~\mbox{\ref{tab:rdr2_tasks}} lists the full task
set. We keep the same adaptive dual controller framework, Event Trigger signal
design, and reporting setup, replacing only the game adapter and task set.

Table~\mbox{\ref{tab:generalization_results}} shows that \method{} reaches
56.4\% success (22/39), compared with 43.6\% for CRADLE and 33.3\% for the
\stardojo{} baseline. This is a 12.8 percentage point gain over CRADLE, with
48.5\% fewer tokens per step and 47.1\% lower latency. The result suggests that
the benefit is not limited to Lite-100's visual style: allocating high-cost
reasoning to event boundaries and supporting those calls with Hierarchical
Memory also transfers to a more visually dynamic game.

\FloatBarrier
\endgroup

%% file: table/main_results.tex
\begin{table}[H]
    \centering
    \small
    \caption{\textbf{Main comparison on \stardojo{}.} ``Recovery/Stuck Ratio'' is
    Recovery Rate divided by Stuck Rate.
    Entries report mean $\pm$ standard deviation over three repeated
    runs. 
    Table~\mbox{\ref{tab:raw_recovery_events}} reports the raw stuck and
    recovery statistics behind this ratio. Bold and underlined entries denote
    the best and second-best results within each method group;
    blue shading marks results from our method.}
    \label{tab:main_results}
    \resizebox{\linewidth}{!}{%
    \begin{tabular}{lccccc}
        \toprule
        Method & Lite-100 SR $\uparrow$ & Budgeted SR $\uparrow$ & Tokens / Task (k) $\downarrow$ & Latency / Step (s) $\downarrow$ & Recovery/Stuck Ratio $\uparrow$ \\
        \midrule
        ReAct-like~\citep{yao2023react} & {\normalsize \phantom{0}6.7\std{0.6\%}} & {\normalsize \phantom{0}9.6\std{0.6\%}} & {\normalsize \best{102.6\std{5.4}}} & {\normalsize \best{24.5\std{3.2}}} & {\normalsize 1.61\std{0.11}} \\
        Reflexion-like~\citep{shinn2023reflexion} & {\normalsize \phantom{0}8.7\std{1.0\%}} & {\normalsize 10.4\std{0.6\%}} & {\normalsize 178.5\std{6.6}} & {\normalsize \second{35.9\std{3.8}}} & {\normalsize 2.02\std{0.14}} \\
        Voyager-like~\citep{wang2024voyager} & {\normalsize \phantom{0}7.3\std{0.6\%}} & {\normalsize \phantom{0}8.1\std{1.0\%}} & {\normalsize 238.3\std{6.7}} & {\normalsize 56.5\std{6.0}} & {\normalsize 1.73\std{0.15}} \\
        CRADLE~\citep{tan2024cradle} & {\normalsize \second{13.0\std{1.7\%}}} & {\normalsize 10.7\std{0.6\%}} & {\normalsize 372.5\std{8.8}} & {\normalsize 73.5\std{8.2}} & {\normalsize \second{2.49\std{0.18}}} \\
        \stardojo{}~\citep{tan2025stardojo}  & {\normalsize 12.3\std{1.4\%}} & {\normalsize \second{12.3\std{1.4\%}}} & {\normalsize 295.9\std{8.4}} & {\normalsize 48.7\std{4.5}} & {\normalsize 2.37\std{0.18}} \\
        \rowcolor{oursrowblue}
        \best{\method{} (Default Qwen3.5-397B)} & {\normalsize \best{18.0\std{1.7\%}}} & {\normalsize \best{21.6\std{1.7\%}}} & {\normalsize \second{168.1\std{5.7}}} & {\normalsize 43.5\std{4.7}} & {\normalsize \best{2.75\std{0.12}}} \\
        \hline
        \rowcolor{oursrowblue}
        \best{\method{} (GPT-5.4)} & {\normalsize \best{20.3\std{1.7\%}}} & {\normalsize \best{23.7\std{1.4\%}}} & {\normalsize \best{107.8\std{2.1}}} & {\normalsize \best{27.9\std{4.3}}} & {\normalsize \best{3.50\std{0.29}}} \\
        \rowcolor{oursrowblue}
        \best{\method{} (Gemini-3.1-pro)} & {\normalsize \second{17.7\std{1.4\%}}} & {\normalsize \second{20.3\std{1.4\%}}} & {\normalsize \second{134.3\std{3.2}}} & {\normalsize \second{33.8\std{3.3}}} & {\normalsize \second{2.66\std{0.19}}} \\
        \bottomrule
    \end{tabular}
    }
\end{table}

%% file: table/task_breakdown.tex
\begin{table}[H]
    \centering
    \small
    \caption{\textbf{Task-type success breakdown on Lite-100.} Each column reports the
    success rate for one task family. Entries report mean $\pm$ standard deviation over
    three repeated runs.}
    \label{tab:task_breakdown}
    \resizebox{\linewidth}{!}{%
    \begin{tabular}{lcccccc}
        \toprule
        Model & Farming $\uparrow$ & Crafting $\uparrow$ & Exploration $\uparrow$ & Combat $\uparrow$ & Social $\uparrow$ & Total $\uparrow$ \\
        \midrule
        ReAct-like~\citep{yao2023react} & 14.3\std{0.0\%} & \phantom{0}9.5\std{4.1\%} & \phantom{0}4.8\std{2.1\%} & 0.0\std{0.0\%} & \phantom{0}4.0\std{0.0\%} & \phantom{0}6.7\std{0.6\%} \\
        Reflexion-like~\citep{shinn2023reflexion} & 17.5\std{2.7\%} & 11.9\std{4.1\%} & \phantom{0}4.8\std{2.1\%} & 2.8\std{2.8\%} & \phantom{0}6.7\std{2.3\%} & \phantom{0}8.7\std{0.6\%} \\
        Voyager-like~\citep{wang2024voyager} & 14.3\std{0.0\%} & \phantom{0}9.5\std{4.1\%} & \phantom{0}6.0\std{2.1\%} & 2.8\std{2.8\%} & \phantom{0}4.0\std{0.0\%} & \phantom{0}7.3\std{0.6\%} \\
        CRADLE~\citep{tan2024cradle}  & \second{25.4\std{2.7\%}} & 16.7\std{4.1\%} & \second{10.7\std{3.6\%}} & 2.8\std{2.8\%} & \phantom{0}\second{8.0\std{4.0\%}} & \second{13.0\std{1.0\%}} \\
        \stardojo{}~\citep{tan2025stardojo}  & 20.6\std{2.7\%} & \second{19.0\std{4.1\%}} & \phantom{0}9.5\std{2.1\%} & \second{5.6\std{4.8\%}} & \phantom{0}\second{8.0\std{0.0\%}} & 12.3\std{0.6\%} \\
        \hline

        \rowcolor{oursrowblue}
        \best{\method{} (Ours)} & \best{34.9\std{2.7\%}} & \best{23.8\std{4.1\%}} & \best{13.1\std{2.1\%}} & \best{8.3\std{8.3\%}} & \best{10.7\std{2.3\%}} & \best{18.0\std{1.7\%}} \\
        \bottomrule
    \end{tabular}
    }
\end{table}

%% file: table/scheduler_sensitivity.tex
\begin{table}[t!]
    \centering
    \small
    \caption{\textbf{Event Trigger threshold sensitivity on Lite-100.}}
    \label{tab:scheduler_sensitivity}
    \resizebox{\linewidth}{!}{%
    \begin{tabular}{lccp{0.32\linewidth}cccc}
        \toprule
        Setting & $T$ & $\tau_v$ & Stall trigger & Lite-100 SR  &
        Budgeted SR  & Tokens / Task (k)  &
        Strategic calls / step  \\
        \midrule
        More strategic & 3 & 0.30 &
        $z_t\ge3$ or $(r_t\ge4 \wedge z_t\ge2)$ &
        17.0\% & 20.4\% & 163.3 & 2.355 \\
        Default & 4 & 0.35 &
        $z_t\ge4$ or $(r_t\ge5 \wedge z_t\ge2)$ &
        18.0\% & 21.6\% & 168.1 & 2.273 \\
        More reactive & 6 & 0.40 &
        $z_t\ge5$ or $(r_t\ge5 \wedge z_t\ge3)$ &
        15.4\% & 19.4\% & 172.5 & 2.189 \\
        No periodic refresh & $\infty$ & 0.35 &
        $z_t\ge4$ or $(r_t\ge5 \wedge z_t\ge2)$ &
        11.6\% & 13.8\% & 173.8 & 2.024 \\
        \bottomrule
    \end{tabular}
    }
\end{table}

%% file: table/ablation_results.tex
\begin{table}[H]
    \centering
    \small
    \caption{\textbf{Framework-level and component ablations of \method{} on Lite-100.}
    ``$\Delta$ SR'' is the absolute change in
    Lite-100 SR in percentage points relative to the full \method{}.
    Entries report mean $\pm$ standard deviation over three repeated runs.}
    \label{tab:ablation_results}
    \resizebox{\linewidth}{!}{%
    \begin{tabular}{lccccc}
        \toprule
        Variant & Lite-100 SR  & Tokens / Task (k)  & Latency / Step  & Recovery/Stuck Ratio  & $\Delta$ SR \\
        \midrule
        \multicolumn{6}{l}{\emph{Framework-level ladder}} \\
        \rowcolor{oursrowblue}
        \best{\method{} (Full)} & {\normalsize \best{18.0\std{1.7\%}}} & {\normalsize 168.1\std{5.7}} & {\normalsize 43.5\std{4.7}} & {\normalsize \best{2.75\std{0.12}}} & {\normalsize \best{0.0}} \\

        Single Strategic Controller + Global Memory & {\normalsize 12.7\std{1.0\%}} & {\normalsize 286.8\std{7.3}} & {\normalsize 57.8\std{5.4}} & {\normalsize 2.18\std{0.09}} & {\normalsize -5.3} \\

        Single Strategic Controller + Hierarchical Memory & {\normalsize \second{16.3\std{0.6\%}}} & {\normalsize 305.4\std{6.9}} & {\normalsize 61.1\std{5.0}} & {\normalsize \second{2.32\std{0.14}}} & {\normalsize \second{-1.9}} \\

        Single Reactive Controller + Global Memory & {\normalsize \phantom{0}6.7\std{0.6\%}} & {\normalsize \best{116.8\std{7.3}}} & {\normalsize \best{28.8\std{3.4}}} & {\normalsize 1.84\std{0.05}} & \phantom{0}{\normalsize -11.3} \\

        Single Reactive Controller + Hierarchical Memory & {\normalsize 10.3\std{0.6\%}} & {\normalsize \second{135.4\std{6.9}}} & {\normalsize \second{33.1\std{3.9}}} & {\normalsize 2.05\std{0.08}} & {\normalsize -7.7} \\

        Dual Controller + Global Memory & {\normalsize 12.3\std{1.0\%}} & {\normalsize 154.5\std{4.9}} & {\normalsize 39.8\std{4.1}} & {\normalsize 1.98\std{0.12}} & {\normalsize -5.7} \\
        \midrule
        \multicolumn{6}{l}{\emph{Component removals}} \\
        w/o Event Trigger & {\normalsize 15.7\std{0.6\%}} & {\normalsize 198.5\std{8.3}} & {\normalsize 39.4\std{5.8}} & {\normalsize 2.01\std{0.12}} & {\normalsize -2.3} \\

        w/o reactive override & {\normalsize 15.7\std{0.6\%}} & {\normalsize 184.7\std{6.5}} & {\normalsize 49.4\std{4.8}} & {\normalsize 1.97\std{0.08}} & {\normalsize -2.3} \\

        w/o \samb{} & {\normalsize 15.0\std{1.0\%}} & {\normalsize 160.3\std{6.1}} & {\normalsize 42.1\std{4.6}} & {\normalsize 2.10\std{0.12}} & {\normalsize -3.0} \\

        w/o SA-KG & {\normalsize 14.3\std{0.6\%}} & {\normalsize 158.9\std{5.2}} & {\normalsize 40.8\std{4.4}} & {\normalsize 1.96\std{0.12}} & {\normalsize -3.7} \\

        \bottomrule
    \end{tabular}
    }
\end{table}

%% file: table/raw_recovery_events.tex
\begin{table}[t!]
    \centering
    \small
    \caption{\textbf{Raw stuck and recovery statistics on Lite-100.} ``Stuck Events''
    counts move-blocked and empty-action events from the execution logs.
    ``Recovered Events'' counts stuck events after which the agent returns to a
    normal task-execution state within the next three environment steps. Raw
    fields are populated from the same runs used for
    Table~\mbox{\ref{tab:main_results}}; auxiliary model rows follow Appendix
    Sec.~\mbox{\ref{sec:appendix_model_token_note}}.}
    \label{tab:raw_recovery_events}
    \resizebox{\linewidth}{!}{%
    \begin{tabular}{lrrrrccc}
        \toprule
        Method & Total Steps & Stuck Events & Recovered Events & Stuck Rate $\downarrow$ & Recovery Rate $\uparrow$ & Recovery/Stuck Ratio $\uparrow$ \\
        \midrule
        ReAct-like~\citep{yao2023react} & 5875\std{31} & 2115\std{16} & \best{1227\std{9}}\phantom{0} & 36.0\% & 58.0\% & 1.61\std{0.11} \\
        Reflexion-like~\citep{shinn2023reflexion} & 5820\std{22} & 1804\std{13} & 1129\std{8}\phantom{0} & 31.0\% & 62.6\% & 2.02\std{0.14} \\
        Voyager-like~\citep{wang2024voyager} & 5842\std{46} & 1957\std{22} & \second{1135\std{10}} & 33.5\% & 58.0\% & 1.73\std{0.15} \\
        CRADLE~\citep{tan2024cradle} & \second{5445\std{64}} & \second{1497\std{27}} & 1026\std{15} & \best{27.5\%} & \second{68.5\%} & \second{2.49\std{0.18}} \\
        \stardojo{}~\citep{tan2025stardojo} & 5536\std{40} & 1550\std{15} & 1029\std{8}\phantom{0} & 28.0\% & 66.4\% & 2.37\std{0.18} \\
        \rowcolor{oursrowblue}
        \method{} (Ours) & \best{5195\std{43}} & \best{1434\std{22}} & 1089\std{12} & \second{27.6\%} & \best{75.9\%} & \best{2.75\std{0.12}} \\
        \hline
        \rowcolor{oursrowblue}
        \method{} (GPT-5.4) & \best{4630\std{25}} & \best{912\std{11}} & \second{629\std{6}}\phantom{0} & \best{19.7\%} & \second{69.0\%} & \best{3.50\std{0.29}} \\
        \rowcolor{oursrowblue}
        \method{} (Gemini-3.1-pro) & \second{5215\std{38}} & \second{1413\std{20}} & \best{1022\std{14}} & \second{27.1\%} & \best{72.3\%} & \second{2.66\std{0.19}} \\
        \bottomrule
    \end{tabular}
    }
\end{table}

%% file: table/generalization_results.tex
\begin{wraptable}{r}{0.56\linewidth}
    \vspace{-0.9em}
    \centering
    \scriptsize
    \caption{\textbf{Cross-game generalization on Red Dead Redemption 2 (RDR2).} SR is
    reported as mean $\pm$ standard deviation over three repeated runs on 13 tasks.}
    \label{tab:generalization_results}
    \resizebox{\linewidth}{!}{%
    \begin{tabular}{lccc}
      \toprule
      Method & SR  & Tokens / Step  & Latency / Step (s)  \\
      \midrule
      ReAct-like~\citep{yao2023react} & 15.4\std{2.6\%} & \best{2.1k\std{0.2k}} & \phantom{0}\best{28.2\std{2.7}} \\
      Reflexion-like~\citep{shinn2023reflexion} & 17.9\std{5.1\%} & 3.5k\std{0.2k} & \phantom{0}\second{41.4\std{3.6}} \\
      Voyager-like~\citep{wang2024voyager} & \phantom{0}7.7\std{2.6\%} & 4.8k\std{0.3k} & \phantom{0}72.4\std{3.9} \\
      CRADLE~\citep{tan2024cradle} & \second{43.6\std{7.7\%}} & 6.6k\std{0.8k} & 102.2\std{8.4} \\
      \stardojo{}~\citep{tan2025stardojo} & 33.3\std{5.1\%} & 5.7k\std{0.5k} & \phantom{0}67.5\std{4.1} \\
      \rowcolor{oursrowblue}
      \best{\method{} (Ours)} & \best{56.4\std{7.7\%}} & \second{3.4k\std{0.7k}} & \phantom{0}54.1\std{4.6} \\
      \bottomrule
    \end{tabular}
    }
    \vspace{-1.0em}
\end{wraptable}

%% file: sec/5_conclusion.tex
\section{Conclusion}
\label{sec:conclusion}

This paper shows that long-horizon game agents benefit from coupling structured
control with structured memory. \textbf{\method{}} uses adaptive dual control to reserve
expensive strategic reasoning for planning and recovery, while leaving routine
execution to a lightweight reactive controller. Together with Hierarchical
Memory, this design improves success, cost, and recovery on Lite-100 and also
transfers to RDR2 under different visual-domain and action-interface assumptions.
Ablations further show that \textbf{event triggering}, \textbf{reactive
override}, and \textbf{controller-specific memory} are complementary rather than
interchangeable parts of the framework.

\paragraph{Limitations and future work.}
\label{sec:limitations}
\method{} can still misroute control when progress is visually subtle, failure
signals are noisy, when a game state changes without large screen differences,
or when retrieved memory becomes stale. Its evidence is strongest on Lite-100
and selected RDR2 tasks, so broader game styles, lower-level interfaces, and
longer deployments remain important future tests. The current system uses visual
observations and action history, while audio cues and richer multimodal feedback
are not modeled. Although \method{} reduces unnecessary large-model calls,
recovery and replanning still depend on explicit model invocations, leaving cost
and latency as practical constraints for deployment. Future work should learn
event-triggering and memory-writing policies from interaction data, add
uncertainty-aware retrieval, incorporate additional task signals, and study
whether training or demonstrations can improve the lightweight controller
without sacrificing adaptability.

%% file: sec/X_suppl.tex
\phantomsection
\section*{\LARGE Appendix}
\label{appendix}

\subsection*{Contents}

\newcommand{\appcontentssection}[3]{%
  \noindent\makebox[\linewidth]{\textbf{#1\quad #2}\hfill\textbf{\pageref{#3}}}\par\vspace{0.06em}}
\newcommand{\appcontentsitem}[3]{%
  \noindent\hspace*{2.1em}\makebox[\dimexpr\linewidth-2.1em\relax]{#1\quad #2\leaders\hbox to 0.55em{\hss.\hss}\hfill\pageref{#3}}\par\vspace{0em}}

\appcontentssection{A}{Metric and Reproducibility Details in \stardojo{}}{sec:appendix_reproducibility}
\appcontentsitem{A.1}{Step-budget SR versus Budgeted SR}{sec:appendix_step_budget_vs_budgeted}
\appcontentsitem{A.2}{Compute Resources}{sec:appendix_compute_resources}
\appcontentsitem{A.3}{Additional Reproducibility Settings}{sec:appendix_additional_reproducibility_settings}
\appcontentsitem{A.4}{Auxiliary Model Rows}{sec:appendix_model_token_note}
\vspace{0.16em}
\appcontentssection{B}{Baseline and Protocol Details}{sec:appendix_protocol_details}
\appcontentsitem{B.1}{Baseline Adaptation}{sec:appendix_baseline_adaptation}
\appcontentsitem{B.2}{LLM-call Accounting}{sec:appendix_llm_call_accounting}
\appcontentsitem{B.3}{Recovery-event Counting}{sec:appendix_recovery_event_counting}
\appcontentsitem{B.4}{RDR2 Protocol and Task List}{sec:appendix_rdr2_protocol}
\appcontentsitem{B.5}{Demo Visualizations}{sec:appendix_demo_visualizations}
\vspace{0.16em}
\appcontentssection{C}{Controller Prompt Skeletons}{sec:appendix_prompt_templates}
\appcontentsitem{C.1}{Condensed Prompt Skeletons}{tab:prompt_skeletons}
\appcontentsitem{C.2}{Concrete Prompt Excerpts}{sec:appendix_concrete_prompt_excerpts}
\vspace{0.16em}
\appcontentssection{D}{Hyperparameter Summary}{sec:appendix_hyperparameters}
\appcontentsitem{D.1}{Hyperparameter Table}{tab:hyperparameter_summary}
\vspace{0.16em}
\appcontentssection{E}{Event Trigger Implementation Details}{sec:appendix_scheduler_details}
\appcontentsitem{E.1}{Signal Definitions}{sec:appendix_scheduler_signals}
\appcontentsitem{E.2}{Component-ablation Interpretation}{sec:appendix_component_ablation_interpretation}
\vspace{0.16em}
\appcontentssection{F}{Hierarchical Memory Implementation Details}{sec:appendix_memory_details}
\appcontentsitem{F.1}{Memory Ranking and Fusion}{sec:appendix_memory_ranking_fusion}
\appcontentsitem{F.2}{Assets and Licenses}{sec:appendix_assets_licenses}
\vspace{0.16em}
\appcontentssection{G}{Additional Limitations}{sec:appendix_limitations}
\appcontentsitem{G.1}{Task-family Disparities}{sec:appendix_task_family_disparities}
\appcontentsitem{G.2}{Representative Combat Failure Case}{sec:appendix_representative_combat_failure}
\vspace{0.16em}
\appcontentssection{H}{Responsible Use and LLM Disclosure}{sec:appendix_responsible_use}
\appcontentsitem{H.1}{Ethics Statement}{sec:appendix_ethics_statement}
\appcontentsitem{H.2}{Reproducibility Statement}{sec:appendix_release_reproducibility_statement}
\appcontentsitem{H.3}{Broader Social Impact}{sec:appendix_broader_social_impact}
\appcontentsitem{H.4}{LLM Disclosure}{sec:appendix_llm_disclosure}

\section{Metric and Reproducibility Details in \stardojo{}}
\label{sec:appendix_reproducibility}

For reproducibility, this appendix fixes the task split, evaluation budgets,
model backend, LLM-call accounting rule, visual encoder access, and the
run-level logging needed to reconstruct each reported metric.
All quantitative results in the main paper report mean and standard deviation
over three repeated runs. Both Lite-100 SR and Budgeted SR are evaluated on the
\stardojo{} Lite-100 benchmark split, which contains 100 tasks covering farming,
crafting, exploration, combat, and social interaction, with a difficulty split
of 56 easy, 23 medium, and 21 hard tasks. Lite-100 SR is the original
\stardojo{} success rate under the benchmark step caps. Budgeted SR keeps the
same tasks and difficulty split, but replaces the step cap with the equivalent
large-model call budget derived from the benchmark-native \stardojo{} protocol.

Specifically, the \stardojo{} baseline invokes the large model four times per
environment step, and Lite-100 uses step caps of 30, 50, and 150 for easy,
medium, and hard tasks. Multiplying these step caps by four gives the Budgeted
SR call budgets of 120, 200, and 600 calls for the three difficulty levels. One
LLM call means one invocation of the
Qwen3.5-397B-A17B backend that returns one model completion, whether it is made
by the Strategic Controller or the Reactive Controller. Calls used for
reflection, reasoning, action proposal, reactive action selection, memory
querying, retrieval-query summarization, retry, or replanning are all counted
separately. Visual encoding, deterministic vector or index lookup, Event Trigger
rules, and environment stepping are not counted as LLM calls because
they do not return a model completion. Token usage is reported separately as
tokens per task.

Unless otherwise stated, both controllers in \method{} use the same base model,
Qwen3.5-397B-A17B. The Strategic Controller uses a careful long-context
configuration, whereas the Reactive Controller uses a token-constrained
direct-action configuration. All compared agents use the same LLM backend,
Qwen3.5-397B-A17B, and share the same environment interface, action space,
prompt upper bound, and decoding settings. The same frozen visual encoder is
available to methods that include a visual change or visual retrieval
component; methods without such a component do not call it. Because Budgeted SR
changes the stopping rule from a fixed environment step cap to a fixed
large-model call budget, methods that use fewer large-model calls per step can
execute more environment actions before exhausting the budget. Budgeted SR can
therefore be either higher or lower than standard Lite-100 SR, and should be
read as an efficiency metric under equal LLM call budgets rather than a stricter
subset of the standard evaluation.

\paragraph{Step-budget SR versus Budgeted SR.}
\label{sec:appendix_step_budget_vs_budgeted}
The two SR metrics answer different questions. Lite-100 SR follows the original
\stardojo{} step budget and measures completion under fixed interaction
opportunities, but it does not penalize agents that spend more LLM calls inside
each step and can favor high-call-per-step methods. Budgeted SR fixes the
large-model call budget instead, making LLM-call efficiency visible, but it has
the opposite bias: low-call methods may receive more environment steps before
their budget is exhausted. We therefore report both metrics together with token
usage and latency. A baseline's lower Budgeted SR can reflect early exhaustion
of the artificial call budget rather than an inherent inability to solve the
task.

\paragraph{Compute resources.}
\label{sec:appendix_compute_resources}
The shared Qwen3.5-397B-A17B backend is served on two eight-GPU MI308X nodes,
while CPU workers handle environment execution, logging, memory indexing, and
result aggregation. Local environment execution, artifact checks, and
aggregation were run on a workstation with 64 GB system memory and an NVIDIA
RTX 4060 Ti GPU with 16 GB memory. Experiments are run with eight concurrent
rollouts. Each reported number is calculated from three repeated Lite-100 runs.
We report end-to-end wall-clock latency per step in the main tables because
this metric includes event triggering, memory retrieval, model inference, action-return
overhead, and possible communication delay between environment workers and the
model-serving nodes. Such transmission delay may affect the measured latency,
although all methods use the same serving and concurrency setup. The final
reported runs will be accompanied by the corresponding worker configuration,
memory budget, and run-time logs.

\paragraph{Additional reproducibility settings.}
\label{sec:appendix_additional_reproducibility_settings}
The Lite-100 runner uses eight parallel
workers by default and sets \texttt{FIXED\_SEED=true} and
\texttt{PYTHONHASHSEED=42}; the framework default fixed seed is 42. Unless
overridden by an environment variable, the default decoding temperature is 1.0;
fixed-seed Strategic Controller runs use temperature 0.0, and the Reactive
Controller's direct-action request uses temperature 0.1 with thinking disabled.
Stage-specific output limits are 1024 tokens for image or state description,
800 for task inference, 512 for self-reflection, and 2048 for action planning;
the Reactive Controller uses a 1024-token maximum. LLM-side image inputs are
capped at 960-pixel resolution.
The recent-history window is five steps. The local image-change encoder is
\texttt{local-image-embedding-v1} with 1024 dimensions. The State-Action Memory
Bank (\samb{}) quick-path retrieval threshold is 0.85, the execution threshold
is 0.92, and the returned \samb{} hint count is one. The State-Action Knowledge
Graph (SA-KG) uses BAAI/bge-base-en-v1.5 embeddings with 768
dimensions, ChromaDB storage, a maximum of 10,000 entries, top-$k=5$, a
similarity threshold of 0.85, and write filtering that requires reflection
success with confidence at least 0.7.

The local image-change encoder is a deterministic visual descriptor rather than
a pretrained external checkpoint. Each observation is EXIF-normalized, converted
to grayscale, resized to $32 \times 32$ with bilinear interpolation, flattened
to 1024 dimensions, scaled to $[0,1]$, mean-centered, and L2-normalized before
cosine distance is computed. This encoder is used only for visual change
detection; it is separate from the SA-KG text embedding model
\texttt{BAAI/bge-base-en-v1.5}.

\paragraph{Auxiliary model rows.}
\label{sec:appendix_model_token_note}
The GPT-5.4 and Gemini-3.1-pro rows in Table~\mbox{\ref{tab:main_results}} are
reported as controlled auxiliary backend runs: they use the same Lite-100 split,
call budgets, \method{} configuration, and evaluation protocol while swapping
the large-model backend. Tokens per task still depend on model verbosity, retry
behavior, recovery frequency, and Event Trigger calls, so the gap between
GPT-5.4 and Gemini-3.1-pro reflects the full execution trace rather than model
quality alone. Their latency is further affected by serving
infrastructure: these closed-source backends are accessed through commercial
model-serving APIs, whereas Qwen3.5-397B-A17B is self-hosted on our MI308X
nodes, so faster calls can partly reflect provider-side serving optimization.

\section{Baseline and Protocol Details}
\label{sec:appendix_protocol_details}

This section spells out the shared evaluation interface used for all compared
agents. The goal is to isolate controller allocation and memory design rather
than differences in backend access, action parsing, retry policy, or hidden
environment state. All methods therefore receive the same task text, screenshot
or image-derived observations, executable action schema, recent execution
feedback, and bounded history summary. Method-specific additions such as
reflection, memory, or skill reuse are allowed only when they are part of that
method's intended decision pattern, and they are built from the method's own
rollout history.

\paragraph{Baseline adaptation.}
\label{sec:appendix_baseline_adaptation}
All baselines are evaluated through the same \stardojo{} runner, action space,
task success evaluator, maximum prompt budget, decoding settings, and
Qwen3.5-397B-A17B backend used by \method{}. We keep each baseline's core
decision pattern while standardizing only the input/output interface needed for
Lite-100 execution. Their prompts are filled from the same runtime fields
whenever applicable: task description, current observation or image-derived
facts, executable action schema, skill library, previous action and execution
feedback, and the current history summary. The current task, observation,
action schema, and latest feedback are kept as required fields; older history
and method-specific memory fields are bounded by the same recent-history window
of five steps and the same maximum prompt budget.

ReAct-like keeps a reason-act loop over the shared action API and uses only the
rolling interaction history. Reflexion-like adds verbal failure reflection and
feeds the reflection back into later prompts from its own rollout history.
Voyager-like keeps a memory/skill reuse interface, but stores only skills or
summaries produced in the \stardojo{} runs and executes the same primitive
actions as other methods. CRADLE keeps its perception-planning-execution
workflow under the shared backend and environment interface. The benchmark-native
\stardojo{} row uses the original four-call-per-step protocol from \stardojo{}
under the same runner and evaluation script. LLM transport retries,
short-response retries, and parser or format retries use the shared runner
configuration; any repeated large-model invocation is counted by the Budgeted
SR call counter. Method-specific memories or reflections are isolated and are
built only from each method's own rollout history.

The shared runner validates every proposed action against the same executable
schema before dispatch. Invalid or unparsable outputs trigger the same retry
path across methods and are still counted as large-model completions when they
require another model request. Episode termination is also shared: a rollout ends
when the benchmark success evaluator accepts the task, the task reaches its
configured step or call budget, or the runner encounters a terminal environment
failure. These rules prevent a baseline from gaining extra interaction time
through parser differences or uncounted recovery calls.

\paragraph{LLM-call accounting.}
\label{sec:appendix_llm_call_accounting}
Budgeted SR counts every large-model completion request made by an agent,
including first-pass calls, transport retries, short-response retries, parser
or format retries, reflection calls, task-reasoning calls, action-proposal
calls, reactive action-selection calls, memory-query calls, retrieval-query
summarization calls, retry calls, and replanning calls. Any LLM completion
invoked for summarization, reflection, reasoning, memory querying, or action
selection is counted. Visual encoding, deterministic vector or index lookup,
Event Trigger rules, and environment stepping remain outside the
LLM-call counter because they do not return a model completion.

\paragraph{Recovery-event counting.}
\label{sec:appendix_recovery_event_counting}
Let $N_{\mathrm{step}}$ be the number of environment steps,
$N_{\mathrm{stuck}}$ the number of move-blocked or empty-action events recorded
in the execution log, and $N_{\mathrm{rec}}$ the number of such events followed
by a return to normal task execution within the next three environment steps. We
report
\begin{equation}
\label{eq:recovery_metrics}
\mathrm{StuckRate}=N_{\mathrm{stuck}}/N_{\mathrm{step}},\quad
\mathrm{RecoveryRate}=N_{\mathrm{rec}}/\max(1,N_{\mathrm{stuck}}),
\end{equation}
and the Recovery/Stuck Ratio is
$\mathrm{RecoveryRate}/\max(\mathrm{StuckRate},\epsilon)$ with a small
$\epsilon$ only to avoid division by zero. Table~\mbox{\ref{tab:raw_recovery_events}}
reports the raw fields behind this metric. Rates and ratios are computed for
each run and then averaged across the three repeated runs; the reported
uncertainty is the sample standard deviation across those runs. To keep the raw
event table internally consistent with the reported mean rates and mean total
steps, event-count means are rounded from the corresponding mean rate and mean
step values, while the event-count standard deviations are empirical across
runs.
The ratio is intended to diagnose recovery behavior conditional on stuck
events. If $N_{\mathrm{stuck}}=0$, the run has no recovery opportunity and
should be reported as ``No stuck events'' rather than interpreted as a zero
Recovery/Stuck Ratio.

\paragraph{RDR2 protocol.}
\label{sec:appendix_rdr2_protocol}
The RDR2 generalization experiment follows the CRADLE benchmark setup: all
methods use the same environment configuration, task interface, action
space, and success-evaluation protocol. The experiment uses 13 tasks, each run
three times, for 39 total rollouts. We keep the same controller allocation and
Event Trigger signal design as in the Lite-100 evaluation while replacing only
the game environment adapter and task set. For each repeated run, we compute the
aggregate success rate over the 13 tasks, then report the mean and sample
standard deviation across the three repeated runs; the standard deviation is not
computed over the pooled 39 task-level rollouts.

\paragraph{RDR2 task list.}
Table~\mbox{\ref{tab:rdr2_tasks}} lists the 13 tasks used in the RDR2
generalization experiment. Each task is run three times, yielding 39 total
rollouts.

\begin{table}[H]
    \centering
    \small
    \caption{\textbf{RDR2 task list for cross-benchmark generalization.}}
    \label{tab:rdr2_tasks}
    \begin{tabular}{rlrl}
        \toprule
        \# & Task & \# & Task \\
        \midrule
        1 & Follow Dutch & 8 & Lead Horse \\
        2 & Go to Town & 9 & Follow Javier \\
        3 & Hitch Horse & 10 & Search John \\
        4 & Protect Dutch & 11 & Keep Wolves Away \\
        5 & Search for Supplies & 12 & Kill Wolves \\
        6 & Go to Barn & 13 & Return to Camp \\
        7 & Search Barn & & \\
        \bottomrule
    \end{tabular}
\end{table}

\begin{figure}[H]
    \centering
    \includegraphics[width=1\linewidth]{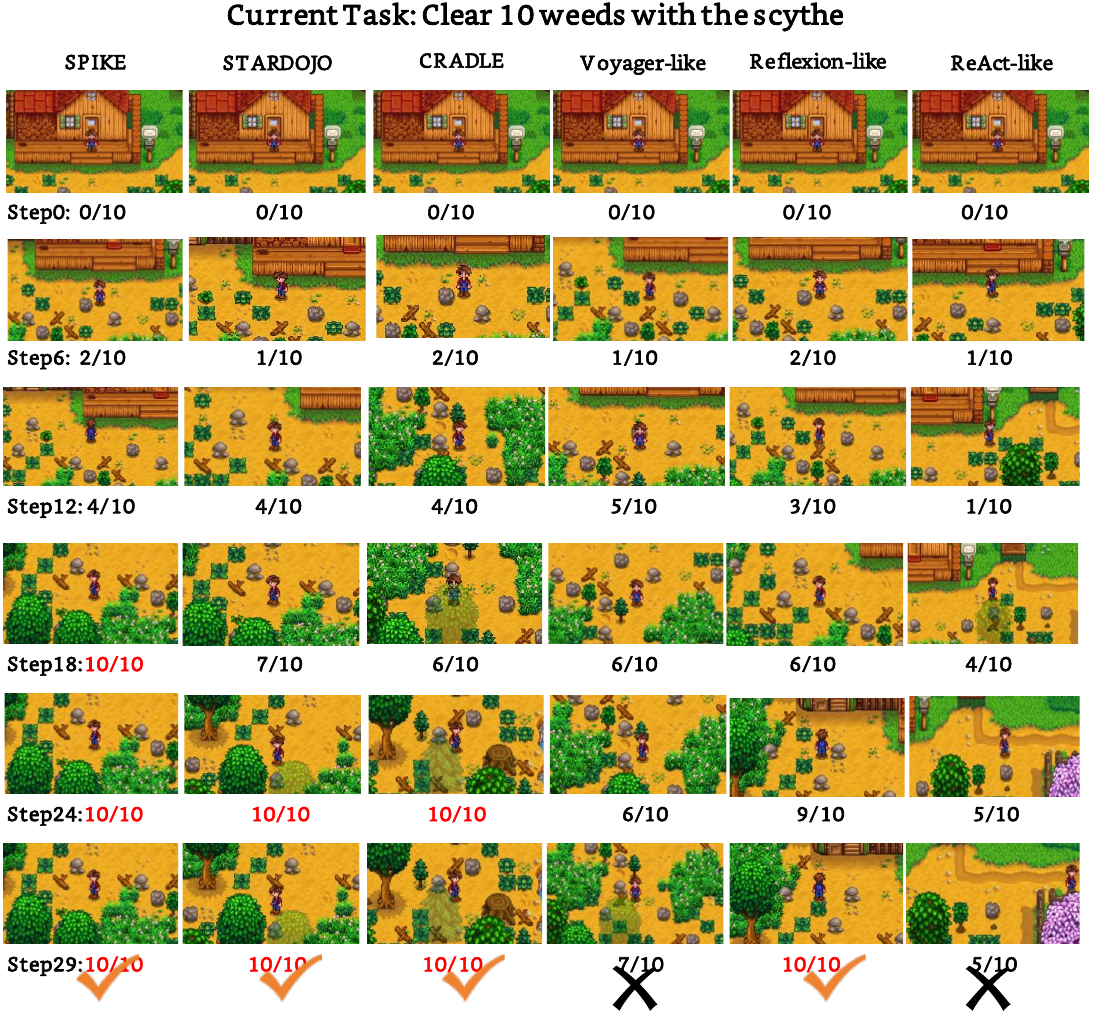}
    \caption{\textbf{\stardojo{} demonstration example.} The task is to clear
    10 weeds with the scythe. The annotated progress counters show how each
    agent advances over time; \method{} reaches the target earlier and maintains
    stable progress under the same visual-control interface.}
    \label{fig:appendix_stardojo_demo}
\end{figure}

\begin{figure}[H]
    \centering
    \includegraphics[width=0.94\linewidth]{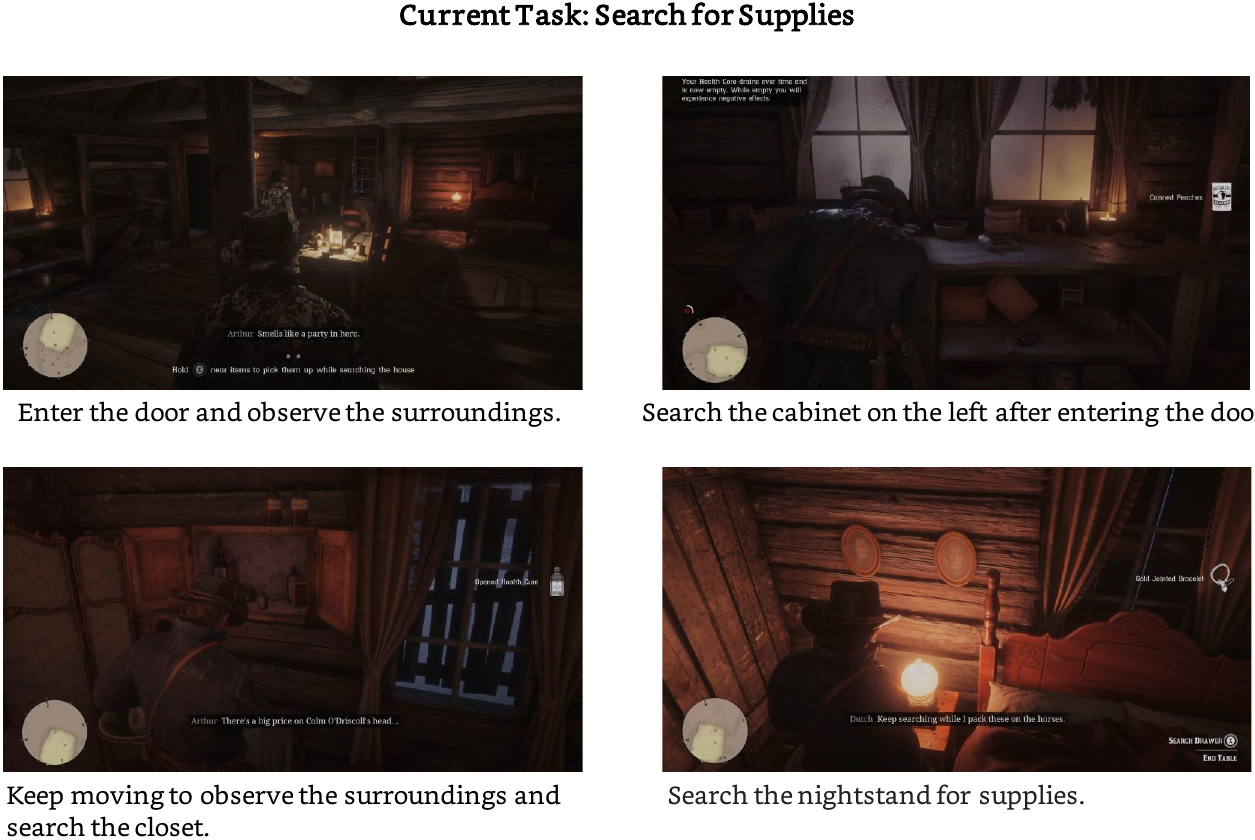}
    \caption{\textbf{RDR2 demonstration example.} The task is to search for
    supplies. The annotations mark the high-level plan sequence: enter the
    building, inspect the room, search the cabinet and closet, and check the
    nightstand, illustrating transfer to a visually different navigation and
    search setting.}
    \label{fig:appendix_rdr2_demo}
\end{figure}

\paragraph{Demo visualizations.}
\label{sec:appendix_demo_visualizations}
Figures~\mbox{\ref{fig:appendix_stardojo_demo}} and
\mbox{\ref{fig:appendix_rdr2_demo}} provide qualitative examples from the
\stardojo{} and RDR2 settings, respectively.
Figure~\mbox{\ref{fig:appendix_rdr2_demo}} shows transfer to a visually different
3D setting, where search is decomposed into room entry, object inspection, and
local interaction rather than relying on \stardojo{}-specific screen patterns.
Figure~\mbox{\ref{fig:appendix_stardojo_demo}} shows the complementary
\stardojo{} behavior: \method{} keeps progress stable on a tool-use task while
reserving Strategic Controller calls for stalls or scene changes.

\section{Controller Prompt Skeletons}
\label{sec:appendix_prompt_templates}

Table~\mbox{\ref{tab:prompt_skeletons}} summarizes the prompt skeletons used by
the main controller stages; task-specific templates use the same fields with
domain-specific constraints such as shopping, cultivation, combat, or navigation.
The table is a condensed excerpt rather than a full prompt dump: runtime fields
such as observations, SA-MB hints, SA-KG evidence, controller state, and output
schemas are filled by the runner, and the complete templates will be released
with the implementation.

\newcommand{\promptsnippet}[2]{%
\begin{center}
\begin{tabular}{|p{0.95\linewidth}|}
\hline
\rowcolor{black!75}\textcolor{white}{\bfseries #1}\\
\hline
\begin{minipage}{0.93\linewidth}
\vspace{0.3em}
{\ttfamily\scriptsize\raggedright #2}
\vspace{0.3em}
\end{minipage}\\
\hline
\end{tabular}
\end{center}
}

\begin{table}[H]
    \centering
    \small
    \caption{\textbf{Condensed prompt skeletons for the controller stages.}}
    \label{tab:prompt_skeletons}
    \resizebox{\linewidth}{!}{%
    \begin{tabular}{p{0.20\linewidth}p{0.43\linewidth}p{0.29\linewidth}}
        \toprule
        Stage & Main inputs & Required output \\
        \midrule
        Information collection &
        Screenshot description, on-screen text, UI state, item status, location,
        dialogue, and environment facts. &
        Categorized state summary used by both controllers. \\
        Self Reflection &
        Current task/subtask, previous action and reasoning, current and previous
        toolbar state, recent history, visual context, and environment facts. &
        Action-effect judgment, likely failure cause, next-step suggestion, and
        subtask/task completion status. \\
        Strategic Controller &
        Task, subtask, acquisition target or route hint, game/UI state,
        surroundings, blocker summaries, recent feedback, memory evidence, and
        valid action set. &
        Short reasoning plus a bounded next-action proposal for the Reactive
        Controller. \\
        Reactive Controller &
        Latest Strategic Controller proposal, current observation, \samb{}
        memory hints, execution feedback, recent action log, and valid action
        set. &
        One grounded executable action in the shared action API, or escalation
        only when no grounded action is available. \\
        \bottomrule
    \end{tabular}
    }
\end{table}

\paragraph{Concrete prompt excerpts.}
\label{sec:appendix_concrete_prompt_excerpts}
The following boxes show shortened excerpts from the actual prompt templates;
placeholders are filled at runtime from the current game state, retrieved memory,
and execution logs.
They are organized to preserve the closed-loop contract used by \method{}:
the Strategic Controller proposes a bounded local intent, the Reactive Controller
grounds that intent in the latest observation and may only make local corrections,
and Self Reflection converts execution feedback into progress or failure signals
for the next Event Trigger decision.

\promptsnippet{SPIKE Strategic Controller Prompt Template}{%
You are a helpful AI assistant integrated with `Stardew Valley' on the PC,
equipped to handle various tasks in the game.\par
Here is some helpful information to help you make the decision.\par
Current task: \detokenize{<$task_description$>}\par
Current subtask: \detokenize{<$subtask_description$>}\par
Current acquisition target item: \detokenize{<$target_item$>}\par
Location / time / season / health / energy / money:
\detokenize{<$location$>}, \detokenize{<$time$>}, \detokenize{<$season$>},
\detokenize{<$health$>}, \detokenize{<$energy$>}, \detokenize{<$money$>}.\par
Current position and facing:
\detokenize{<$current_position$>}, \detokenize{<$facing_direction$>}.\par
Surroundings, front-tile summary, blocker signature, nearest grounded target,
inventory, toolbar, buildings, NPCs, exits, recent feedback, and valid action
set are then provided.\par
Event Trigger state: \detokenize{<$trigger_reason$>},
\detokenize{<$zero_progress_streak$>}, \detokenize{<$repeat_tail$>},
\detokenize{<$failure_level$>}.\par
SA-KG evidence for plan checking:
\detokenize{<$retrieved_state_action_graph$>}.\par
Output format: brief situation assessment; next local subgoal;\par
candidate actions from the valid action set;\par
stop condition for returning control to the Event Trigger.%
}

\promptsnippet{SPIKE Reactive Controller Prompt Template}{%
You are a helpful AI assistant integrated with `Stardew Valley' on the PC,
serving as the execution brain. The planning brain has already suggested an
action. Your job is to compare that suggestion with the latest actual game state
and choose the single best action to execute right now.\par
Current task: \detokenize{<$task_description$>}\par
Current subtask: \detokenize{<$subtask_description$>}\par
Current observation: location / time / season / health / energy / money:
\detokenize{<$location$>}, \detokenize{<$time$>}, \detokenize{<$season$>},
\detokenize{<$health$>}, \detokenize{<$energy$>}, \detokenize{<$money$>}.\par
Position, facing direction, menu state, inventory, selected item, surroundings,
front-tile summary, toolbar, and latest visual context:
\detokenize{<$current_position$>}, \detokenize{<$facing_direction$>},
\detokenize{<$current_menu$>}, \detokenize{<$inventory$>},
\detokenize{<$chosen_item$>}, \detokenize{<$surroundings$>},
\detokenize{<$front_tile_summary$>}, \detokenize{<$toolbar_information$>},
\detokenize{<$image_introduction$>}.\par
SA-MB memory hints and recent state summary:
\detokenize{<$history_summary$>}\par
Execution feedback: \detokenize{<$action_feedback$>}\par
Recent execution feedback: \detokenize{<$recent_execution_feedback$>}\par
Planning brain's suggested action: \detokenize{<$suggested_action$>}\par
Reason: \detokenize{<$suggested_reason$>}\par
Execution history in current cycle: \detokenize{<$execution_log$>}\par
Valid action set: \detokenize{<$skill_library$>}\par
If the suggestion matches current facts, adopt it.\par
Otherwise choose a better grounded action, but only as a bounded local
correction such as facing adjustment, obstacle avoidance, tool reselection, or
one-step repositioning.\par
Do not change the current subgoal. If no grounded local action can advance the
subgoal, request escalation instead of inventing a new plan.\par
Output exactly one executable action or an escalation request.%
}

\promptsnippet{SPIKE Self Reflection Prompt Template}{%
Assume you are a helpful AI assistant integrated with `Stardew Valley' on the
PC. Your task is to examine the inputs, interpret the in-game context, and
determine whether the executed action took effect.\par
Current task: \detokenize{<$task_description$>}\par
Current subtask: \detokenize{<$subtask_description$>}\par
Last executed action: \detokenize{<$previous_action$>}\par
Previous reasoning: \detokenize{<$previous_reasoning$>}\par
History summary: \detokenize{<$history_summary$>}\par
Before/after observations, toolbar state, selected item, inventory delta, menu
state, position/facing change, and execution message:
\detokenize{<$state_transition_summary$>}.\par
Answer whether the action succeeded.\par
Identify the most probable failure cause.\par
Suggest the next step.\par
Decide whether the subtask or task is completed.\par
Return a compact status record: success, progress evidence, failure level,\par
and whether the Event Trigger should escalate.%
}

\section{Hyperparameter Summary}
\label{sec:appendix_hyperparameters}

\begin{table}[H]
    \centering
    \small
    \caption{\textbf{Fixed hyperparameters used in the reported experiments.}}
    \label{tab:hyperparameter_summary}
    \resizebox{\linewidth}{!}{%
    \begin{tabular}{lll}
        \toprule
        Component & Parameter & Value \\
        \midrule
        \stardojo{} protocol & Easy/medium/hard step caps & 30 / 50 / 150 \\
        Budgeted SR & Easy/medium/hard LLM-call budgets & 120 / 200 / 600 \\
        Evaluation & Repeated runs per task & 3 \\
        Event Trigger & Periodic strategic refresh $T$ & 4 \\
        Event Trigger & Action-history window $W$ & 5 \\
        Event Trigger & Visual-change threshold $\tau_v$ & 0.35 \\
        Event Trigger & Zero-progress threshold $\tau_z$ & 4 \\
        Event Trigger & Repetition / repeated-stall thresholds $\tau_r / \tau_{rz}$ & 5 / 2 \\
        Event Trigger & Failure threshold $\tau_\ell$ & 2 \\
        \samb{} & Semantic weights $\alpha_C / \alpha_L$ & 0.55 / 0.45 \\
        \samb{} & Reward / success weights $\alpha_R / \alpha_P$ & 0.15 / 0.10 \\
        \samb{} & Reward exponential moving average (EMA) update $\eta$ & 0.3 \\
        \samb{} & Recency decay & 24 hours \\
        SA-KG & Embedding model & BAAI/bge-base-en-v1.5 \\
        SA-KG & Retrieval threshold / top-$k$ $\tau_{\mathrm{KG}} / k_{\mathrm{KG}}$ & 0.85 / 5 \\
        SA-KG & Graph weights $\beta_C / \beta_P$ & 0.6 / 0.4 \\
        Fusion retrieval & Weights $\lambda_{\mathrm{MB}} / \lambda_{\mathrm{KG}}$ & 0.75 / 0.25 \\
        \bottomrule
    \end{tabular}
    }
\end{table}

\section{Event Trigger Implementation Details}
\label{sec:appendix_scheduler_details}

\paragraph{Event-triggered amortized deliberation algorithm.}
\label{sec:appendix_event_algorithm}
Algorithm~\mbox{\ref{alg:event_triggered_deliberation}} summarizes the online
loop used by \method{}. It is written as a reproducibility skeleton: the concrete
threshold values are listed in Table~\mbox{\ref{tab:hyperparameter_summary}}, and
the signal definitions are expanded below.

\begin{figure}[H]
\centering
\small
\fbox{\begin{minipage}{0.94\linewidth}
\textbf{Algorithm A1: Event-triggered amortized deliberation in \method{}}\\
\textbf{Input:} task $x$, thresholds $\Theta$, \samb{}, SA-KG, budget $B$\\
\textbf{Initialize:} proposal $p\leftarrow\emptyset$, history $h_0\leftarrow\emptyset$\\
\textbf{for} $t=1,\ldots,H$ \textbf{do}\\
\quad Observe $o_t=(v_t,u_t,f_t)$ and update history context.\\
\quad Compute visual change $d_t$, stall streak $z_t$, repetition tail $r_t$, and failure level $\ell_t$.\\
\quad Set $g_t\leftarrow\mathrm{EventTrigger}(d_t,z_t,r_t,\ell_t,c_t;\Theta)$.\\
\quad \textbf{if} $g_t=1$ \textbf{then} retrieve SA-KG/global evidence and update proposal $p$ with the Strategic Controller.\\
\quad \textbf{else} retrieve local hints from \samb{}.\\
\quad Emit $a_t$ with the Reactive Controller using $o_t$, $p$, and retrieved hints.\\
\quad Execute $a_t$, receive feedback, and write memory only on success, interpretable recovery, or planner-validated progress.\\
\textbf{end for}
\end{minipage}}
\caption{\textbf{Reproducibility skeleton for the \method{} control loop.}}
\label{alg:event_triggered_deliberation}
\end{figure}

\label{sec:appendix_scheduler_signals}
The fixed thresholds below are not tuned per task; they instantiate
event-boundary detection under the shared \stardojo{} interface and are held
constant across Lite-100 runs unless explicitly varied in the sensitivity study.
Signals instantiate the symbolic thresholds in Eq.~\mbox{\ref{eq:scheduler_routing}}
with the fixed settings in Table~\mbox{\ref{tab:hyperparameter_summary}}. The
Event Trigger records the same-action tail length $r_t$, the zero-progress streak
$z_t$, and the numeric progress $\Delta_t=q_t-q_{t-1}$. In Lite-100, the
progress counter uses only shared runner feedback: task or subgoal completion,
position or facing change, selected-item change, inventory delta, menu/dialogue
transition, and confirmed productive execution. Invalid actions, execution
errors, and repeated productive actions without observable state change do not
increase $q_t$ and instead contribute to $z_t$ or $\ell_t$. These are the same
raw feedback fields exposed by the common wrapper to all methods; the derived
Event Trigger variables are internal to \method{}. The Failure Detector assigns
$\ell_t$ from multimodal traces, with structured execution messages used as
additional evidence.

For visual change detection, consecutive observations $v_{t-1}$ and $v_t$ are
encoded by a frozen visual encoder $\phi(\cdot)$, and the Event Trigger
uses cosine distance:
\begin{equation}
\label{eq:visual_distance}
d_t = 1 - \frac{\phi(v_t)^{\top}\phi(v_{t-1})}{\|\phi(v_t)\|\,\|\phi(v_{t-1})\|}.
\end{equation}
If $d_t > \tau_v$, the event is treated as a scene change, such as an unexpected
enemy, menu pop-up, or map transition, and control is escalated before locally
plausible but globally wrong actions accumulate. Near static visual state plus
repetitive actions contributes to stuck detection, but escalation requires weak
progress as well as repetition.

The failure levels are defined as follows. $\ell_t=0$ denotes normal execution.
$\ell_t=1$ denotes an isolated local issue and triggers bounded local retry.
$\ell_t=2$ denotes hard execution errors, invalid actions, or low progress loops
and escalates with the recent failure trace as negative evidence. $\ell_t=3$
denotes repeated hard failures that invalidate the current plan, flush
misleading short-term context, and force top-level replanning. These
hyperparameters, together with the memory fusion settings, are fixed for all
reported Lite-100 experiments and ablations.

\paragraph{Component-ablation interpretation.}
\label{sec:appendix_component_ablation_interpretation}
Table~\mbox{\ref{tab:ablation_results}} exposes two coupled failure modes.
Removing the Event Trigger spreads Strategic Controller calls across routine steps
and weakens adaptation at state transitions; removing reactive override leaves
local execution errors until the next escalation. The memory removals affect the
evidence available at escalation points: \samb{} mainly supports fast reuse of
recent interaction fragments, while SA-KG supplies structured state evidence for
progress checking and replanning. These patterns support the main-text claim
that scheduling and memory design are coupled rather than independent additions.

\paragraph{Threshold sensitivity.}
\label{sec:appendix_threshold_sensitivity}
Table~\mbox{\ref{tab:scheduler_sensitivity}} in the main text tests whether the
Event Trigger depends on a single hand-tuned threshold choice. The \emph{More
strategic} setting escalates earlier, the \emph{More reactive} setting delays
escalation, and \emph{No periodic refresh} removes the periodic refresh trigger
while leaving event-driven scene-change, stall, and failure triggers active. The
sweep is intended to test the success--cost trend of adaptive allocation rather
than to perform exhaustive hyperparameter search.
The action-history window remains fixed at $W=5$ for all Event Trigger sensitivity
settings; the more-reactive setting delays repeated-action escalation by
requiring a longer zero-progress streak rather than by increasing $W$.

The sweep shows that the default Event Trigger is not merely an arbitrary midpoint.
Making the Event Trigger more strategic increases Strategic Controller calls from
2.273 to 2.355 per step, but slightly lowers Lite-100 SR and Budgeted SR,
suggesting that earlier escalation can interrupt otherwise stable local
execution. Making it more reactive reduces strategic calls to 2.189 per step,
but also lowers success, indicating that delayed escalation leaves stale plans
and local failures active for longer. Removing periodic refresh still leaves
2.024 strategic calls per step under event-driven triggers, but causes the
largest success drop, showing that event-driven triggers alone are insufficient
for long-horizon recovery. Overall, Table~\mbox{\ref{tab:scheduler_sensitivity}}
supports the default design: periodic refresh plus event-driven escalation gives
the best success--cost balance among the tested settings.
Because the non-default rows are lightweight one-run sensitivity checks, this
table is used to inspect local trends rather than to estimate uncertainty.

All sweep settings use the same Lite-100 tasks, backend, prompts, environment
interface, memory configuration, and aggregation scripts as the main runs. The
default row corresponds to the fixed setting used throughout the main
experiments, while each non-default setting is run once as a lightweight
sensitivity check. The Strategic-calls-per-step metric counts large-model
invocations made by Strategic Controller modules and divides them by the number
of environment steps.

\section{Hierarchical Memory Implementation Details}
\label{sec:appendix_memory_details}

\label{sec:appendix_memory_ranking_fusion}
Each \samb{} item stores a state summary, action or short action trace, reward
EMA, empirical success rate, timestamp, and source tag. Repeated writes update
reward EMA as $\mathrm{rewardEMA}_i\leftarrow(1-\eta)\mathrm{rewardEMA}_i+\eta r_t$
with $\eta=0.3$. Before retrieval, query and state text are lowercased,
punctuation is stripped, and whitespace is collapsed. The lexical term is
$L_i=|\mathcal{T}(x_t)\cap\mathcal{T}(s_i)|/
|\mathcal{T}(x_t)\cup\mathcal{T}(s_i)|$, and the cosine term $C_i$ is computed
from normalized token-frequency vectors. Reward, reliability, and recency use
$R_i=(\mathrm{clip}(\mathrm{rewardEMA}_i,-1,1)+1)/2$,
$P_i=\mathrm{successes}_i/\max(1,\mathrm{attempts}_i)$, and
$\rho_i=\exp(-\mathrm{ageHours}_i/24)$. Thus all retrieval terms are on
comparable $[0,1]$ scales before weighting. The Reactive Controller receives
the top $k_{\mathrm{MB}}$ retrieved summaries as hints rather than executable
commands.

SA-KG stores validated transitions $(s_t,a_t,s_{t+1})$ as state nodes and
action edges. Repeated transitions update edge counts, reward EMA, and
reliability. During escalation, the Strategic Controller queries the top five
nearest state nodes with a Chroma vector index built from the
BAAI/bge-base-en-v1.5 embedding model, keeps candidates with
$C_j=1-\mathrm{dist}(x_t,s_j)\ge\tau_{\mathrm{KG}}$, and selects outgoing
actions by success rate with execution count as a tie-breaker.

Fusion retrieval uses the \samb{} ranking as the primary list and injects the
largest matched SA-KG action boost for each candidate action. For each retrieval
call, source scores are normalized within their own candidate sets before
fusion:
\begin{equation}
\label{eq:score_normalization}
\begin{aligned}
\tilde S_{\star}(c)=
\frac{S_{\star}(c)-\min_{u\in\mathcal{C}_{\star}}S_{\star}(u)}
{\max_{u\in\mathcal{C}_{\star}}S_{\star}(u)-
\min_{u\in\mathcal{C}_{\star}}S_{\star}(u)+\epsilon},
\quad \star\in\{\mathrm{MB},\mathrm{KG}\}.
\end{aligned}
\end{equation}
The final fused score is
\begin{equation}
\label{eq:fused_retrieval_score}
\begin{aligned}
S_{\mathrm{fuse}}=\lambda_{\mathrm{MB}}\tilde S_{\mathrm{MB}}+\lambda_{\mathrm{KG}}\tilde S_{\mathrm{KG}},
\end{aligned}
\end{equation}
where $S_{\mathrm{KG}}$ follows Eq.~\mbox{\ref{eq:sakg_score}} and unmatched graph actions use
$\tilde S_{\mathrm{KG}}=0$. After execution, successful actions update \samb{}
statistics immediately. A trajectory is promoted to SA-KG only after successful
completion, interpretable recovery, or Strategic Controller validation, which
prevents transient errors from becoming long-term evidence.

\paragraph{Assets and licenses.}
\label{sec:appendix_assets_licenses}
The work uses existing research assets rather than introducing a new dataset or
new foundation model. We cite the benchmark, prior agent methods, memory
methods, and graph-memory work used for comparison or motivation in the main
paper. The released materials will document external implementation assets,
including benchmark assets, model interfaces, evaluation scripts, and
figure-generation dependencies, together with corresponding license or usage
terms when available.

\section{Additional Limitations}
\label{sec:appendix_limitations}

The proposed framework introduces more moving parts than a single-controller
baseline, including Event Trigger thresholds, retrieval quality, and write filtering.
These components create new failure modes: stale memory may be retrieved, local
control may remain active too long, and escalation may happen either too late or
too often. Strong performance on \stardojo{} also does not automatically imply
transfer to other games or embodied settings, especially when their action
abstractions or observation formats differ greatly.

The recovery metric uses benchmark-provided execution logs to make stuck and
recovery events comparable across methods. These logs are not required by the
online Event Trigger, but deployment settings with only screenshots and
low-level inputs would force F1--F3 detection to rely more on visual
differencing, action-history patterns, and inferred progress. Such settings may
make failure grading noisier, especially for subtle blocked movements or menu
states that are visually unclear.

The remaining limitations separate category-specific bottlenecks from one
representative failure trace, and then return to attribution limits for the
multi-component design.

\paragraph{Task-family disparities.}
\label{sec:appendix_task_family_disparities}
The lower Combat and Social success rates in Table~\mbox{\ref{tab:task_breakdown}}
reflect category-specific bottlenecks rather than a uniform controller failure.
Combat tasks require both finding the target enemy and reacting while monsters
continue moving, so visual change and local navigation errors can invalidate a
plan within a few steps. Social tasks often require traveling from the player's
home to another character's home over long routes before the interaction can
begin, placing heavier demands on navigation, route persistence, and subgoal
planning than shorter farming or crafting tasks.

\begin{figure}[t]
    \centering
    \includegraphics[width=0.92\linewidth]{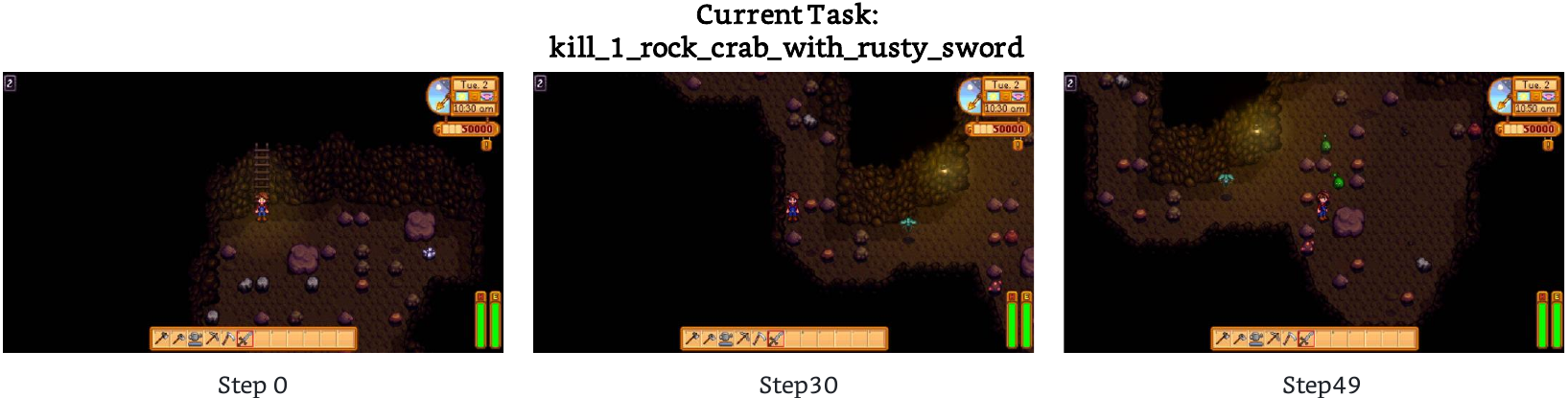}
    \caption{\textbf{Representative combat failure case.} The triptych shows a
    rock-crab task in the mine: the agent begins with the correct combat tool
    and search objective, repeatedly navigates through cluttered rocks while the
    target remains hard to localize, and eventually exhausts the step budget
    without task progress.}
    \label{fig:appendix_failure_case}
\end{figure}

\paragraph{Representative combat failure case.}
\label{sec:appendix_representative_combat_failure}
A representative failure occurs in a rock-crab combat task
(Figure~\mbox{\ref{fig:appendix_failure_case}}). The agent selects the weapon
and keeps the high-level objective of searching the mine floor, but then spends
most of its budget on short local moves around obstacles while task progress
remains zero. Replanning events still tend to produce local navigation commands
rather than a reliable search-and-engage policy. This illustrates why Combat
remains difficult: the agent must couple target localization, navigation through
cluttered mine geometry, and attack timing against enemies that may move or
remain visually ambiguous.

Another limitation concerns attribution. Once multiple memory layers are
available, it becomes easy to attribute gains too broadly to memory alone. We
therefore separate the effects of event triggering, summarization, and long-term
retrieval through explicit ablations, but finer-grained causal analysis remains
future work.

\section{Responsible Use and LLM Disclosure}
\label{sec:appendix_responsible_use}

\paragraph{Ethics statement.}
\label{sec:appendix_ethics_statement}
The proposed framework is evaluated only in benchmarked game environments and
does not involve human subjects, crowdsourcing, personally identifiable
information, or private user data. It does not introduce a new external dataset
or a new foundation model. Existing benchmark assets, model interfaces, and
implementation dependencies are cited in the paper. The work does not require
IRB approval because it does not involve human-subject research or interaction
with private user data.

\paragraph{Reproducibility statement.}
\label{sec:appendix_release_reproducibility_statement}
To facilitate reproducibility, we will provide data, code, and instructions with
the final paper, including the controller implementation, Event Trigger
and memory configuration, prompt templates, task split metadata, evaluation
scripts, aggregation scripts, and instructions for reproducing the reported
Lite-100 and RDR2 experiments. The final release will also record the run
configurations and logs used for the reported tables. We do not release a new
foundation-model checkpoint; access to the LLM backend follows the corresponding
model and provider terms.

\paragraph{Broader social impact.}
\label{sec:appendix_broader_social_impact}
Potential positive impacts include more efficient long-horizon assistive agents
for complex games and software-like environments, lower evaluation cost for
research on token--latency--success trade-offs, and reusable tools for studying
recovery behavior in interactive agents. Potential negative impacts include
misuse for game botting, automation that violates platform rules, or transfer to
software-control settings where incorrect actions could cause unintended
effects. Long-term memory may also preserve stale or biased state descriptions
if the underlying observations or model outputs are unreliable. We mitigate
release risk by presenting SPIKE as benchmark-oriented evaluation work rather
than a deployment-ready automation product, and recommend human oversight,
domain restrictions, logging, rate limits, and platform-rule compliance for any
downstream use.

\paragraph{LLM disclosure.}
\label{sec:appendix_llm_disclosure}
LLMs are a core method component of the work rather than a writing aid: the
Strategic Controller and Reactive Controller are both implemented with
large-model calls, as described in the methodology and experimental setup
sections.